\documentclass[journal]{IEEEtran}
\usepackage{algorithm}
\usepackage{algorithmicx}
\usepackage{algpseudocode}
\usepackage{amsmath}
\usepackage{mathrsfs}
\usepackage{amsfonts}
\usepackage[numbers,sort&compress]{natbib}
\usepackage{amssymb}
\usepackage{bm}
\usepackage{enumerate}
\usepackage[pdftex]{graphicx}
\usepackage{lineno}
\usepackage{url}
\emergencystretch=\maxdimen
\ifCLASSINFOpdf
\else
\fi

\begin{document}
%
\title{Deep Optimized Multiple Description Image Coding via Scalar Quantization Learning}
%
%
%

\author{Lijun~Zhao,
        Huihui~Bai,~\IEEEmembership{Member,~IEEE,}
        Anhong~Wang,~\IEEEmembership{Member,~IEEE,}
        and Yao~Zhao,~\IEEEmembership{Senior~Member,~IEEE}

\thanks{Corresponding author: Lijun Zhao and Huihui Bai. This work was supported in part by Doctoral Scientific Research Starting Foundation of Taiyuan University of Science and Technology (No. 20192023), Funding Awards for Outstanding Doctors Volunteering to Work in Shanxi Province (No. 20192055), National Natural Science Foundation of China (No. 61672373) and Key Innovation Team of Shanxi 1331 Project (KITSX1331).}
\thanks{L. Zhao and A. Wang are with the Institute of Digital Media \& Communication, Taiyuan University of Science and Technology, Taiyuan, 030024, P. R. China, e-mail: leejun@tyust.edu.cn}
\thanks{H. Bai and Y. Zhao are with the Beijing Key Laboratory of Advanced Information Science and Network Technology, Institute of Information Science, Beijing Jiaotong University, Beijing, 100044, P. R. China, e-mail: {15112084, hhbai, yzhao}@bjtu.edu.cn.}

}
%
%

\markboth{Journal of \LaTeX\ Class Files}
{Shell \MakeLowercase{\textit{et al.}}: Bare Demo of IEEEtran.cls for IEEE Journals}
%



\maketitle

\begin{abstract}
In this paper, we introduce a deep multiple description coding (MDC) framework optimized by minimizing multiple description (MD) compressive loss. First, MD multi-scale-dilated encoder network generates multiple description tensors, which are discretized by scalar quantizers, while these quantized tensors are decompressed by MD cascaded-ResBlock decoder networks. To greatly reduce the total amount of artificial neural network parameters, an auto-encoder network composed of these two types of network is designed as a symmetrical parameter sharing structure. Second, this autoencoder network and a pair of scalar quantizers are simultaneously learned in an end-to-end self-supervised way. Third, considering the variation in the image spatial distribution, each scalar quantizer is accompanied by an importance-indicator map to generate MD tensors, rather than using direct quantization. Fourth, we introduce the multiple description structural similarity distance loss, which implicitly regularizes the diversified multiple description generations, to explicitly supervise multiple description diversified decoding in addition to MD reconstruction loss. Finally, we demonstrate that our MDC framework performs better than several state-of-the-art MDC approaches regarding image coding efficiency when tested on several commonly available datasets.

\end{abstract}


\IEEEpeerreviewmaketitle
\section{Introduction}
%
%
%
%
\IEEEPARstart{P} {ACKET LOSS} and bit errors may often unavoidably occur when Internet data are transmitted over unreliable channels \cite{qq2}. As increasing more people surf the Internet in daily life with mobile devices, such as hand-held PAD and cell-phone, whose data packets are conveyed over wireless communication channels, incomplete decoding or complete loss of data very possibly occur. Multiple description coding (MDC) \cite{qq2, qq11, qq12} is one of the representative promising mechanisms among different error resilient coding techniques to make real-time transmission systems more simple and robust to a lossy channel in a challenging environment. Compared with layered-based coding, this mechanism does not need to design the priority of each packet or data re-transmission mechanism and does not require any signal feedback. Several independent yet correlated bit streams generated by the MDC technique facilitate to the independent decoding of each packet, while joint decoding of two or more bit-streams is also supported by the MDC technique.

Traditional MDC approaches have been widely studied in the last decades, among which the derivation of multiple description (MD) theoretical rate-distortion regions is a fundamental and significant topic \cite{qq4}. Meanwhile, in practice, the achievable rate-distortion regions gradually approach the boundaries of theoretical MDC rate-distortion regions \cite{qq4, qq11, qq27, qq28, qq35, qq36, qq37, qq12, qq3, qq19}. However, a large number of traditional MDC methods face many difficult problems. For example, MD quantizers often must assign the optimal index for multiple description generations \cite{qq27, qq28, qq17, qq13, qq10, qq26,qq14}, especially when quantizing more than two descriptions, which is an extremely complicated problem. For the correlating transform-based MDC framework, good performance can be achieved for multiple description coding when adding a small amount of redundancy \cite{qq35,qq36,qq37,qq38}. However, this framework does not always perform well if more redundancy is introduced into multiple descriptions. Different from multiple description quantizers and the
correlated transform-based MDC framework, a class of sampling-based multiple description coding is more flexible and compatible with the standard coders. However, most of existing sampling-based MDC methods are built on the specifically designed sampling methods or extend the existing sampling operator for multiple description generations \cite{qq12, qq3, qq19,qq8}, whose coding efficiency is limited. Consequently, the research of sampling-based MDC methods should be further developed. Recently, a convolutional neural network(CNN)-based JPEG-compliant MDC framework \cite{qq11} has been used to sample an input image to adaptively create multiple description images, but its coding efficiency is not very high because the usage of the standard JPEG limits the performance of this framework. In summary, we should comprehensively study the topic of multiple description coding for error resilience against bit errors and packet loss over an unpredictable channel.

In this paper, a deep multiple description image coding framework is proposed for robust transmission. Our contributions are listed below:
\begin{enumerate}[(1)]
\item We design a general deep optimized MD coding framework based on artificial neural networks. Our framework has several main parts: an MD multi-scale-dilated encoder network, a pair of multiple description scalar quantizers, and MD cascaded-ResBlock decoder networks, as well as conditional probability models.
\item A pair of scalar quantization operators is automatically learned in an end-to-end self-supervised way for generation of diversified multiple descriptions. Meanwhile, each scalar quantizer is accompanied by an importance-indicator map to quantize feature tensors according to the change in the image spatial content.
\item We propose the use of the multiple description structural similarity distance loss to supervise the decoded multiple description images, which implicitly regularizes diversified multiple description generations and scalar quantization learning. Please note that our multiple description structural similarity distance loss is different from the pixel-wise mean-absolute-deviation distance loss from \cite{qq11}. These two types of distance loss work on different spaces, among which the proposed distance is imposed on the decoded images, whereas the distance loss from \cite{qq11} regularizes feature tensors produced by the MD generation neural network.
\item A symmetrical parameter sharing structure is designed for our autoencoder network to greatly reduce the total amount of neural network parameters. Specifically, the parameters of the preceding convolutional layers in the encoder network are shared for multiple description generation, while the decoder networks have a symmetrical structure by sharing the parameters of the back layers.
\item A pair of conditional probability models is learned to estimate the informative amount of the quantized tensors, which can supervise the learning of the multiple description multi-scale-dilated encoder network to compactly represent the input image.

\end{enumerate}

The rest of this paper is organized as follows. First, four kinds of MDC methods are reviewed in Section II and the problem formulation of deep image compression is presented in Section III. Second, the proposed multiple description coding framework is introduced in Section IV. Third, experimental results and analysis are presented in Section V. Finally, we conclude our paper in Section VI.

\section{Related works}
We mainly review four kinds of MDC methods: MD quantizers, correlating transform-based MDC methods, sampling-based MDC and standard-compliant MDC.

\subsection{Multiple description quantizers}
For quantization-based MDC methods, there are three primary classes: scalar quantizers, trellis-coded quantizers, and lattice vector quantizers. In early development, MD scalar quantizers constrained by the symmetric entropy are formed as an optimization problem \cite{qq27}. Following this work, linear joint decoders are developed to resolve the problem of dynastical computation increase \cite{qq17} when generalizing this method from two to L descriptions during encoder optimization. In \cite{qq28}, MD distortion-rate performance is derived for certain randomly generated quantizers. By generalizing randomly generated quantizers, the theoretical performance of MDC with randomly offset quantizers is given in the closed-form expressions \cite{qq13}. To increase the robustness to bit errors, linear permutation pairs are developed for index assignment for two description scalar quantizers \cite{qq10}. In \cite{qq26}, a trellis is formed by the tensor product of trellises for multiple description coding.

The scalar quantizers and trellis-coded quantizers usually require the complicated index assignment. As compared with these quantizers, lattice vector quantizers have been widely studied, since lattice vector quantizers have many advantages such as a symmetrical structure, avoidance of complex nearest neighbor search, lack of the need for codebook design, etc.  In \cite{qq14}, the design problem of lattice vector quantizer is cast as a labeling problem, and a systematic construction method is presented for the general lattices. To further improve the performance of MD vector quantization at the cost of a slight complexity increase \cite{qq15}, the fine lattice codebook is replaced by a non-lattice codebook. In \cite{qq18}, a structured bit-error resilient mapping is leveraged to make MD lattice vector quantizers resilient to the transmission bit error by exploiting intrinsic structural characteristic of the lattice. In \cite{qq7}, a heuristic index assignment algorithm is given to control coding distortions in order to balance the reconstruction quality of different descriptions. In summary, MD quantizers always involve complicated index assignments, especially when more than two descriptions are expected to be generated by users, whose complexity is always very high.

\subsection{Correlating transform-based MDC methods}
The transform-based MDC framework employs a pairwise correlating transform to correlate pairs of transform coefficients \cite{qq35}, in which the best strategy for image compression redundancy allocation is given. This framework considers only the case of coding two multiple descriptions. In \cite{qq36}, this transform-based approach is generalized to create L descriptions to enhance the transmission system robustness with a small quantity of redundancy. To satisfy the channel bandwidth ratio, a ratio configurable MD correlating transform coding \cite{qq37} is introduced to adjust the description data size ratio. In \cite{qq38}, a gradient search is provided to determine the correlating transform, while statistical dependencies between different descriptions benefit estimation of the loss transform coefficients under a lossy network during transmission. Although these correlating transform-based MDC methods provide an effective MD generation method, this kind of MDC method tends to be inefficient when a great deal of information redundancy is expected to be introduced into multiple descriptions.

\subsection{Sampling-based multiple description coding} 
Multiple descriptions can be created from different domains for input images/videos coding in various ways. For instance, diverse descriptions can be obtained from the spatial domain, frequency domain or temporal domain. In \cite{qq30}, two MD video coding coders employ a poly-phase downsampling technique to create multiple descriptions. In \cite{qq29}, adaptive temporal-spatial error concealment is applied for MD video coding, in which multiple descriptions are obtained by spatial subsampling. For MD video coding, motion information from temporal domain is often estimated in the encoder as a redundancy. To quickly recover error on the decoder side without complicated motion searching for MD video coding, motion vectors are sampled according to each description's redundancy, which could further strengthen error resilience when video streams are transmitted over error-prone wireless networks \cite{qq31}. By selecting the temporal level key pictures, hierarchical-B-pictures-based MD video coding framework adopts the simplest odd/even frame splitting method of generating redundancy descriptions as a special case of the framework \cite{qq32}. Among these approaches, some of the sampling-based MDC methods can be compatible with standard image/video compression. However, their sampling operators are manually designed, which limits multiple description coding efficiency, so adaptive sampling method of generating multiple descriptions should be deeply investigated, especially the use of deep learning methods.

\subsection{Standard-compliant multiple description coding}
Research on the standard-compliant MDC framework has becomes increasingly significant due to the wide usage of standard image/video codecs such as JPEG and H.264, which has become essential parts of life in practice. Over the past decades, there have been numerous works regarding standard-compliant MD coding \cite{qq12, qq3, qq19}. For instance, an H.264-compliant MD video coding is achieved by interlacing primary and redundant slices \cite{qq12}, whose rate distortion is optimized in an end-to-end manner. In \cite{qq8}, a hybrid layered MD video coding algorithm uses the H.264/AVC encoder to compress a video at low bit rates as the base layer, and then, four-bit streams are divided into two groups as two description enhancement layers after 3D dual-tree discrete wavelet transform and 3D-SPIHT encoding. For HEVC-compatible MD video coding, the redundancy allocation is assigned based on a visual saliency model \cite{qq3}. Most recently, image representation-based compression with CNNs \cite{qq19} has been extended for multiple description image generation to form a standard-compliant convolutional neural network-based MDC framework \cite{qq11}, which is trainable in an end-to-end fashion. Although great progresses have been achieved for the class of standard-compliant multiple description coding \cite{qq8, qq12, qq3}, their coding efficiency has great potential to be enormously improved by artificial neural networks in various ways such as pre-filtering and post-processing \cite{qq11, qq19, qqAR}.

\section{Problem formulation for deep image compression}
In this paper, deep image compression is defined as a new class of image compression approaches that uses artificial neural networks to learn image nonlinear transform and inverse transform from big-data. This class differs from standard image compression with a classical linear transform such as discrete cosine transform or discrete wavelet transform. Since the proposed framework involves extensive machine learning knowledge, we first introduce some definitions such as concepts and terminology for deep image compression. Then, we introduce and review the deep image compression framework. Finally, the extension of deep image coding to multiple description coding is discussed.

\subsection{Definitions}
Generally, the discrete representation of continuous data can be called data quantization, and this quantization is a form of hard quantization. The derivative of the hard type of quantization function is almost zero everywhere, except at the quantized integer point. If a quantization function is differentiable in the definition domain, then it can be called a soft quantization function. In \cite{qq41}, a soft quantization function is defined based on soft allocation with softmax function. Given a $K$-dimension vector $x=[x_1,...,x_K] \in \mathcal{N}^K$, the softmax function for $x$ can be defined as:
\begin{align}
&\xi(x)= softmax(x)=[\zeta(x_1),...,\zeta(x_K)] \in  \mathcal{N}^K,
\end{align}
\begin{align}
&\zeta(x_i)=\frac{e^{x_i}}{\sum_{k=1}^{K} e^{x_k}},
\label{softmax}
\end{align}
in which $i\in [1, 2, ..., K]$. The elements of center variable vector $\mathcal{C}=[c_1, c_2, ...,c_L]$ is not limited to integers and can also be defined as floating-point numbers, a soft assignment on a variable $z\in \mathcal{R}$ can be defined as: $\xi([-\sigma||z-c_1||^2), ... ,-\sigma||z-c_L||^2]) \in  \mathcal{N}^L$, in which $||\cdot||$ denotes the $L2$-norm. The corresponding hard assignment can be written as: $\hat{\xi}(z):$=${\lim_{\sigma \to \infty}}\xi(z)$,
which finally converges to the nearest center $c_k$ of a vector $\mathcal{C}$ in one-hot encoding, that is to say:
\begin{align}
&{\lim_{\sigma \to \infty}} \xi(z)=\left\{\begin{matrix}1  \quad if(\bar{j}=\mathop{\arg\min}_{j\in [1,...,L]} |z-c_{j}|);\\
0  \quad \quad \quad \quad \quad  otherwise; \quad \quad \quad \quad \quad
\end{matrix}\right.
\end{align}
in which $|z-c_{j}|$ is the absolute value of $z-c_{j}$. Here, the one-hot encoding of $c_k$ refers to a zero-vector along the new dimension with length of $L$, except for its $k-$th element to be 1. Using the above soft assignment, the soft quantization $\tilde{q}$ and hard quantization $\hat{q}$ can be defined respectively as:
\begin{align}
&\tilde{q}(z)\in  \mathcal{N}=\sum_{j=1}^{L}c_j \zeta(-\sigma||z-c_j||^2) \notag \\
&=\mathcal{C}\odot \xi([-\sigma||z-c_1||^2, ... ,-\sigma||z-c_L||^2])
\end{align}
\begin{align}
\hat{q}(z)= {\lim_{\sigma \to \infty}} \tilde{Q}(z)= \mathcal{C}\odot \hat{\xi}(z) =c_e,
\end{align}
in which $\odot$ is element-wise multiplication, $c_e$ is the nearest center to $z$ of the center vector $\mathcal{C}$ and $\sigma$ controls the smoothness strength of the soft-quantization function.

\begin{figure}[t]
\centering
\includegraphics[width=3.3in]{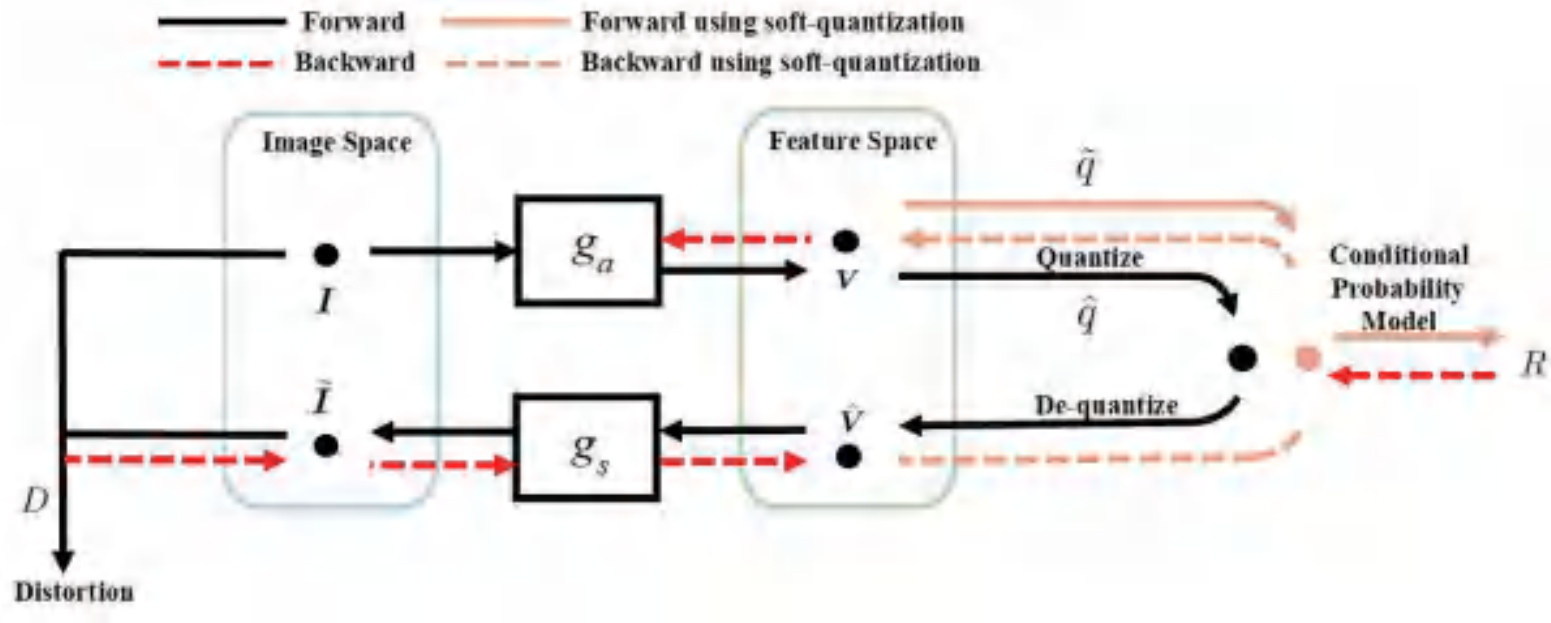}
\caption{The diagram for deep image compression framework.}
\label{DNNMODELa}
\end{figure}

\subsection{Deep image compression framework}
As with the standard coders, general deep image compression usually consists of the following main parts: a nonlinear analysis transform $g_a$ (also known as the encoding network), hard quantization function $\hat{q}$ (as well as soft quantization function $\tilde {q}$ for training), nonlinear synthesis transform $g_s$ (also known as the decoding network), a conditional probability model to estimate the entropy, and distortion function $d(\cdot, \cdot)$ to measure how much image information is lost after compression, as shown in Fig. \ref{DNNMODELa}. Given an input image $\bm{I}$, the feature tensor $\bm{V}$ can be obtained through nonlinear analysis transform $g_a$, i.e., $\bm{V}=g_a(\bm{I})$. Then, the hard quantization function $\hat{q}$ is used to discretize the feature tensor to reduce image compression bits, i.e., $\hat{\bm{V}}=\hat{q}(\bm{V})$. Finally, the conditional probability model $\mathcal{P}_q$ is used to calculate the rate $R(\hat{\bm{V}})$ of the quantized tensor \cite{qq22}. On the decoder side, the quantization vector $\hat{\bm{V}}$ is reconstructed as $\tilde {\bm{I}}$ by the nonlinear synthesis transform $g_s$; that is, $\tilde {\bm{I}}=g_s(\hat{\bm{V}})$. In addition, the distortion function $d(\bm{I},\tilde{\bm{I}})$ is used to measure the distortion of the compressed image. According to the rate-distortion theory of image coding \cite{qq41}, the objective rate-distortion optimization function of deep image compression can be defined as:
\begin{align}
&L(I, \hat{q}, g_a, g_s,\mathcal{P}_q) =\gamma  R(\hat{q}(g_a(\bm{I}))) + D(\bm{I}, g_s(\hat{q}(g_a(\bm{I}))))= \notag \\
& \gamma  E[-log_2 \mathcal{P}_q(\hat{q}(\bm{V}))]+ \{E[d(\bm{I}, g_s(\hat{\bm{V}}))]+ \beta D_{r}(g_a, g_s)\},
\label{objeq}
\end{align}
where the first term is the image compression bit rate constraint, while the second term $D$ includes compressed image distortion $E[d(\bm{I}, g_s(\hat{\bm{V}}))]$ and the network-parameter regularization term $D_{r}(g_a, g_s)$. This equation includes the hard (nondifferentiable) quantization function $\hat{q}$, so it cannot be directly optimized by the general optimization methods, e.g., stochastic gradient descent optimization or Adam optimization. To resolve this problem, a general solution \cite{qq40, qq41, qq42, qq19, qq22} replaces the hard quantization with soft quantization during back-propagation for training (please see the bold dashed orange-pink lines in Fig. \ref{DNNMODELa}), while the forward-propagation uses the hard quantization $\hat{q}$ (please see the bold black lines in Fig. \ref{DNNMODELa}).

The earliest works on image compression based on artificial neural networks primarily study the nondifferentiability problem of the quantization function and the artificial neural network structure design problem \cite{qq41, qq42, qq19, qq22}, as well as how to make compressed images more realistic using the perceptual loss functions \cite{vggloss}. For example, the compressive autoencoder with an encoder network and a decoder network is often chosen as the image compression network \cite{qq40, qq41, qq42, qq19, qq22}, in which the encoder network condenses the input image into a certain amount of bit streams which are as small as possible and are mapped back to a lossy image by the decoder network. In \cite{qq40}, the derivative of the stochastic rounding operation is replaced with the derivative of the expectation for back-propagation, but no part in forward direction is changed. Like \cite{qq40}, soft relaxation of quantization is used to resolve the nondifferentiability problem of the quantization function \cite{qq41}. Different from \cite{qq40, qq41}, the thumbnail images are compressed by a recurrent neural networks architecture, in which a stochastic rounding operation makes feature maps binarized \cite{qq42}. Recently, a virtual codec network has been learned to imitate the projection from the represented vectors to the decoded images to make the image compression framework trainable in an end-to-end way \cite{qq19}. Except for the quantization problem, the autoencoder should be restricted to satisfy the given bit-per-pixel (bpp) compression when learning compressive autoencoder for image compression. In \cite{qq22}, a conditional probability model for the latent distribution of the auto-encoder is learned to supervise the autoencoder network.

\subsection{Extension to deep multiple description image coding}
Until now, there has been no work on deep multiple description coding, although deep image compression methods have been increasingly explored. To the best of our knowledge, our work is the first deep multiple description coding framework. There are two possible methods of deep multiple description coding. In the first, multiple descriptions can be directly generated by an artificial neural network and are discretized by a general quantization to reduce coding bits. If the multiple descriptions refer to multiple description images, which are compressed by the standard codec, then it becomes a multiple description coding framework described in \cite{qq11}. In this framework, JPEG compression can be replaced by a deep image compression approach such as \cite{qq22}. Different from \cite{qq11}, our deep optimized multiple description image coding belongs to the class of methods utilizing the second approach. When the artificial neural network represents an input image as the feature tensors, a pair of scalar quantizers is learned to quantize the feature tensors for diversified multiple description generation. Recently, a convolutional autoencoder-based multiple description coding method \cite{access} has extracts features by learning, which improves image coding efficiency. However, this method suffers from severe coding artifacts similarly to conventional MDC approaches.

\section{The proposed deep multiple description coding}
\begin{figure*}[!htb]
\centering
\includegraphics[width=5.in]{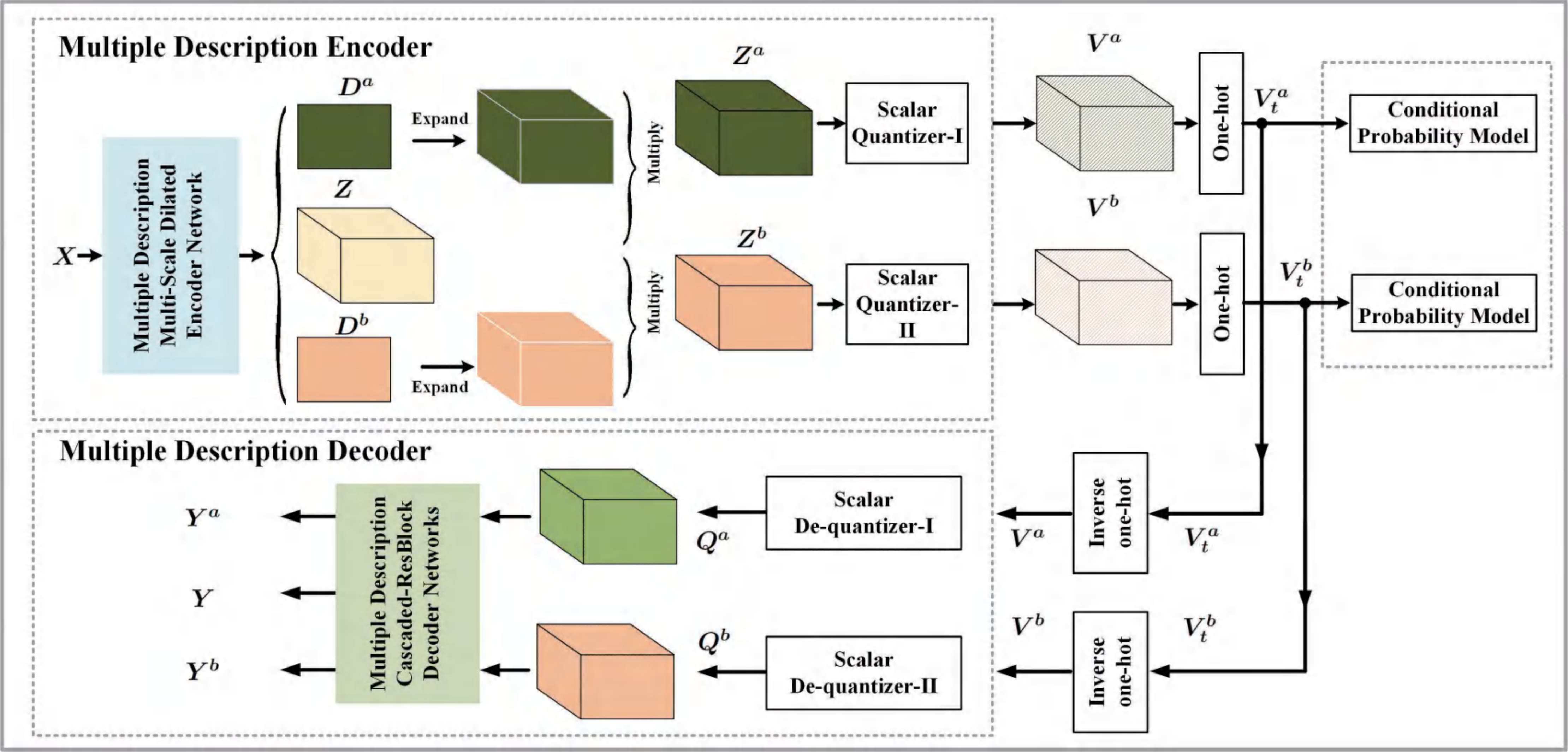}
\caption{The diagram for the proposed deep multiple description coding framework.}
\label{Fig1}
\end{figure*}

In this paper, we introduce a deep optimized multiple description coding framework that is entirely built upon artificial neural networks, as displayed in Fig. \ref{Fig1}. Our framework is primarily composed of a multiple description encoder, a multiple description decoder, and a pair of conditional probability models. The multiple descriptions are generated by the multiple description encoder, which is supervised by the conditional probability models to estimate the entropy of each description, while the received multiple descriptions are decompressed by the multiple description decoder. Specifically, given an input image $\bm{X} \in\mathcal{N}^{(8\times M) \times (8\times N) }$ with a size of $(8\times M) \times (8\times N) $, the multiple description multi-scale-dilated encoder network in the multiple description encoder decomposes the input image into a feature tensor $\bm{Z}\in\mathcal{N}^{M \times N \times K}$, as well as two importance-indicator maps $\bm{D^a}\in\mathcal{N}^{M \times N \times 1}$ and $\bm{D^b}\in\mathcal{N}^{M \times N \times 1}$. Here, $M \times N$ is the spatial size of the feature tensor $\bm{Z}\in\mathcal{N}^{M \times N \times K}$, while $K$ is the number of feature maps for this feature tensor. The importance map indicates which regions should be emphasized during image coding, which is a kind of region-of-interest (ROI) based coding \cite{qqroi}, considering that the spatial distributions of various images are different from each other. M. Li and M. Fabian as well as their co-author, reported that a content-weighted importance map can be used for guidance to allocate image content-aware bit rate for deep image coding based on region-of-interest \cite{qqlimu, qq22}. In consideration of the image spatial variation, each importance-indicator map can be directly multiplied by each feature map of the feature tensor in an element-wise manner; that is, the $K$-feature maps for the feature tensor share the same importance-indicator map. However, vital yet different roles are played by these $K$-feature maps for multiple description image compression. According to the work of M. Fabian and co-workers \cite{qq22}, the expansion operation $\varpi(\bm{D^a})$ for $\bm{D^a}$ is written as:
\begin{align}
&\varpi(\bm{D^a}) \in \mathcal{N}^{K \times M \times N}= \notag\\
&clip(Re(\bm{D^a}*K, [1], [K]) - Re(\mathcal{\aleph}, [2,3], [M, N]), 0, 1),
\label{expanddim}
\end{align}
in which the function $Re(\mathcal{M}, L_{num}, L_{axes})$ is a repetition function used to repeat a given matrix $\mathcal{M}$ with $L_{num}$ times along the axes of $L_{axes}$, to form a new matrix. Furthermore, $[0,...,K-1]\in \mathcal{N}^{K \times 1}$, $\bm{D^a} \in \mathcal{N}^{1 \times M \times N}$, $\mathcal{\aleph}=\varphi([0,...,K-1]^T)\in \mathcal{N}^{K \times 1 \times 1}$, the function $clip(x, 0, 1)$ truncates the value $x$ between $0$ and $1$, and the operation $\varphi$ inserts a dimension of 1 at the last dimension index axis of input shape. We can expand $\bm{D^b}$ in the same way. Then, the expansion of each importance-indicator map is multiplied by the feature tensor $\bm{Z}$ in an element-wise manner to obtain two new feature tensors, $\bm{Z^a}$ and $\bm{Z^b}$, before scalar quantization. From Eq. (\ref{expanddim}), it can be known that only the elements of $\bm{D^a}$ will have valid values for the $i$-th map of $\varpi(\bm{D^a})\in \mathcal{N}^{K \times M \times N}$ along the first dimension, according to the Euclidean distance between $\bm{D^a}*K$ and $\mathcal{\aleph}$. As a result, the $i$-th map of $\varpi(\bm{D^a})$ will influence the multiple description generation of $\bm{Z^a}$. The significance of importance-indicator maps for deep multiple description image coding will be discussed later.

To reduce multiple description coding bits, the two feature tensors $\bm{Z^a}$ and $\bm{Z^b}$ are quantized by scalar quantizer-I and scalar quantizer-II, respectively. Due to the nondifferentiability of hard quantization, we follow the work of E. Agustsson \cite{qq41} to utilize soft quantization to make our framework trainable in an end-to-end way. Scalar quantizer-I with Eq. (\ref{softmax}) is represented as:
\begin{align}
v^a_i=\mathop{\arg\max}_{c_j\in \mathcal{C}^a} \zeta(-\sigma ||z^a(i)-c_j||^2), j \in [1,2,...,n],
\end{align}
where $\mathcal{C}^a=[c_1,...,c_n]$ is a center variable vector and $z^a(i)$ represents the $i$-th element of tensor $\bm{Z^a}$. In a similar way, scalar quantizer-II with a new center variable vector $\mathcal{C}^b$ can be defined. The vectors of the quantized tensors $\bm{V^a}$ and $\bm{V^b}$ can be represented as $[v^a_1,..., v^a_n]$ and $[v^b_1,..., v^b_n]$. This pair of scalar quantizers accompanied by two importance-indicator maps, can be simultaneously learned in an end-to-end self-supervised way. In other words, there is no label for multiple description generation, the ultimate goal of which is to independently decode each side image with an acceptable image quality or jointly decode a better-quality central image, so the side images are used as the opposing labels for each description redundancy measurement with the distance loss for self-supervised learning. Both tensors $\bm{V^a} \in \mathcal{N}^{M \times N \times K}$ and $\bm{V^b}\in \mathcal{N}^{M \times N \times K }$ can be converted into one-hot tensors $\bm{V_t^a}\in \mathcal{N}^{M \times N \times S \times T} $ and $\bm{V_t^b}\in \mathcal{N}^{ M \times N \times S \times T }$ by adding a new dimension, thus producing two descriptions. The generated descriptions are losslessly encoded by arithmetic coding for transmission. During forward-propagation, hard quantization is leveraged to obtain discrete feature tensors according to Eq. (\ref{softmax}), but we use the derivation of the soft quantization function \cite{qq41} to back-propagate gradients from the multiple description decoder to the multiple description encoder, which is conducted in a similar manner as shown in Fig. \ref{DNNMODELa}.

\begin{figure*}[!htb]
\centering
\includegraphics[width=4.3in]{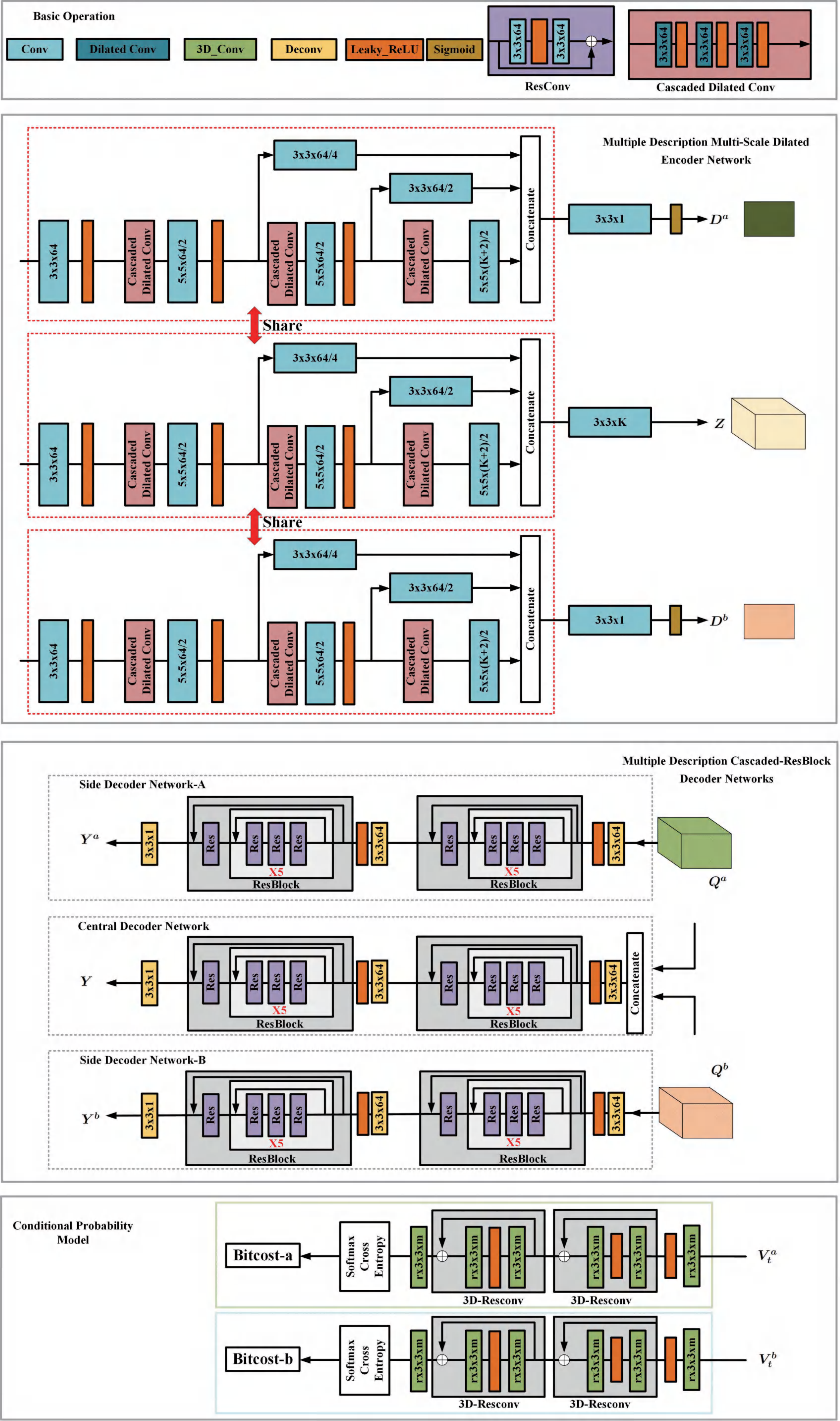}
\caption{The structure diagram of convolutional neural network for deep multiple description coding framework. (Note that 3x3x64 means that spatial kernel size is 3x3, the number of output feature maps is 64, while the stride is 1 in this convolutional layer in default. 3x3x64/2 shares similar denotation except for with a stride of 2. Other convolutional layers can be denoted similarly. X5 in Resblock means that cascaded three-Resconv modules with skip connection are used with five times.)}
\label{MDC-q-network}
\end{figure*}

At the receiver, the decoded one-hot tensors $\bm{V_t^a}$ and $\bm{V_t^b}$ can be reversibly converted into the tensors $\bm{V^a}\in \mathcal{N}^{M\times N\times K}$ and $\bm{V^b}\in \mathcal{N}^{M\times N\times K}$, as displayed in Fig. \ref{DNNMODELa}. Then, these tensors are treated by corresponding scalar de-quantizers. The scalar de-quantizer-I can be written as $\bm{Q^a}(i)= \hbar (\bm{V^a}(i))$, which returns the $\bm{V^a}(i)$-th of the center variable $\bm{C^a}$. Similarly, scalar de-quantizer-II can be written as $\bm{Q^b}(i)= \hbar (\bm{V^b}(i))$. Side decoder network-A (or side decoder network-B) is used to decompress the quantized tensor $\bm{Q^a}$ (or $\bm{Q^b}$) as the lossy image $\bm{Y^a}$ (or $\bm{Y^b}$), as shown in Fig. \ref{MDC-q-network}, if only one description is received at the decoder, when each description is transmitted over an unpredictable channel. If both of the descriptions are received, then the central decoder network is leveraged to jointly decompress quantized tensors $\bm{Q^a}$ and $\bm{Q^b}$ as the central decoded image $\bm{Y}$.

\begin{figure*}[!htb]
\begin{equation}
f_s(\bm{X},\bm{Y})=\Bigg[m \Bigg(\frac{(2 * \mu_{(\bm{X}^5)}\mu_{(\bm{Y}^5)} + c1)}{(\mu_{(\bm{X}^5)}^2 + \mu_{(\bm{Y}^5)}^2 + c1)} \Bigg)\Bigg]^{\omega_5}
*\prod_{i=1}^{4}\Bigg[m \Bigg(\frac{(2 * \sigma_{(\bm{X}^5)}\sigma_{(\bm{Y}^5)} + c2)}{(\sigma_{(\bm{X}^5)}^2 + \sigma_{(\bm{Y}^5)}^2 + c2)} \frac{\sigma_{(\bm{X}^i) (\bm{Y}^i)} + c2}{ \sigma_{(\bm{X}^i)} + \sigma_{(\bm{Y}^i)} + c2}\Bigg)\Bigg]^{\omega^i}
\tag{10}
\label{msssim}
\end{equation}
\hrulefill 
\end{figure*}

\subsection{The objective function of multiple description coding}
Similar to traditional single description image compression, the objective function of multiple description coding requires balancing two fundamental parts: the coding bit rate and multiple description image decoding distortion. The mean square error (MSE) is often employed to measure image compression distortion. However, the human visual perceptual quality of a compressed image with high MSE may be higher than that for images with low MSE \cite{qq20}. There are a multitude of reasons for objective-subjective quality mismatch, such as one-pixel position shifting of the whole-image and new-pixels covered over details-lacking textural regions generated by generative adversarial networks to obtain a sense of reality.

Compared to MSE loss, mean absolute error (MAE) loss can better regularize image compression to advance the compressed images towards the ground-truth images during training \cite{pr2019, pix2pix}. Thus, our framework uses MAE loss for both side decoded images and central decoded image as the first part of our multiple description reconstruction loss, which can be written as follows:
\begin{align}
&D_{1}(\bm{X},\bm{Y^a},\bm{Y^b},\bm{Y})=\frac{1}{64\times M \times N} \cdot \notag\\
&\sum_ {i}(||\bm{X}_i-\bm{Y^a}_i||_1+ ||\bm{X}_i-\bm{Y^b}_i||_1)+\psi*||\bm{X}_i-\bm{Y}_i||_1,
\end{align}
in which $||\cdot||_1$ denotes the $L1$-norm and $\psi$ is a trade-off parameter to control the redundancy between average side reconstructions and the corresponding central reconstruction. Meanwhile, to well measure image distortion for structural preservation, we introduce the multi-resolution (MR) structural similarity index (MR-SSIM) $f_s$ as an evaluation factor of image quality between $\bm{X}$ and $\bm{Y}$ according to \cite{qq21}, which is written in Eq. (\ref{msssim}). Here, $\bm{X}^i$ $(i\in [1, 2, 3, 4, 5])$ is the downsampled image, whose size is $1/[2^{(i-1)}]^2$ times less than that of $\bm{X}$, and $m(\bm{S})$ is the sum operator of $\bm{S}$. Meanwhile, $c1$ and $c2$ are two constants, while $\mu_{(\bm{X}^5)}$ and $\sigma_{(\bm{X}^5)}$ are, respectively, the mean map, the variance map of $\bm{X}^5$ calculated from each pixel's neighborhood window, while $\sigma_{(\bm{X}^5) (\bm{Y}^5)}$ is the covariance map calculated from each pixel's co-located neighborhood windows in $\bm{X}^5$ and $\bm{Y}^5$. Additionally, the weight vector $\omega$ = $[\omega_1, \omega_2, \omega_3, \omega_4, \omega_5]$ for different scales is $[0.750, 0.188, 0.047, 0.012, 0.003]$, which is linearly proportional to each-scale image size; that is, the large-size image weight is greater than that of the small-size image weight. However, image distortions at different scales are of very different importance with respect to perceived quality. In contrast to the MR-SSIM, the MS-SSIM from the literature \cite{qq21} uses the weight vector $\omega^*$ = $[0.0448, 0.2856, 0.3001, 0.2363, 0.1333]$. These weights indicate that the image of other scales are more significant than the largest and smallest scale images. More discussions about MR-SSIM and MS-SSIM will be provided later. The total structural dissimilarity loss as the second part of our multiple description reconstruction loss can be written as:
\begin{equation}
D_{2} (\bm{X}\!, \bm{Y^a}\!, \bm{Y^b}\!, \bm{Y}\!) =- [f_s(\bm{X}\!, \bm{Y^a}\!)+ f_s(\bm{X}\!, \bm{Y^b}\!) + f_s(\bm{X}\!, \bm{Y}\!)].
\tag{11}
\end{equation}

Unlike single description image compression with only one bit stream produced by image coder, multiple description coding should generate diversified multiple descriptions, between which some redundancy is shared, but each description has its unique information. The redundancy of these descriptions makes the receiver capable of decoding an acceptable quality image, even though one description is missing when multiple description bit streams are transmitted over an unstable channel. However, when different descriptions contain excessive shared information, the central image quality will not exhibit great improvements, even though all of the descriptions are obtained by the client. In \cite{qq11}, multiple description distance loss is directly employed to supervise multiple description generation in feature space. Although the feature tensor of each description can be regarded as the opposite label to regularize each other, the learning problem of multiple description coding in feature space is often challenging, because the same multiple description reconstruction may arise from the composition of different features. To allow our framework to automatically generate multiple diverse descriptions, we propose to use the multiple description structural similarity distance loss. This distance loss not only explicitly regularizes decoded multiple description images to be different but also implicitly supervises scalar quantization learning and diversified multiple description generation, which is written as:
\begin{equation}
D_{d}=f_s(\bm{Y^a}, \bm{Y^b}).
\tag{12}
\end{equation}

To compactly represent multiple description feature tensors, there should be an entropy regularization term for our deep MDC framework training. However, our framework is built upon an artificial neural network, so the conditional probability model for the entropy regularization term should be differentiable, i.e., the conditional probability model should also use the artificial neural network. To precisely predict each description coding cost, we use two entropy estimation networks without parameter sharing as the conditional probability model. Following the work of \cite{qq22}, we use context-based entropy estimation neural networks in our framework to efficiently estimate multiple description coding costs during training, as shown in Fig. \ref{Fig1}.

In our framework, the regularization loss from these context-based entropy estimation neural networks supervises the learning of the multiple description encoder, which leads to compact multiple description generation. The estimated coding costs for each description are denoted as $R(\bm{V^a})$ and $R(\bm{V^b})$. Because $\bm{V^a}$ and $\bm{V^b}$ come from the hard quantizers, the $R(\bm{V^a})$ is nondifferentiable, although context-based entropy estimation neural networks are differentiable. If our scalar quantization uses the soft quantizers to obtain $\bm{\tilde{V}^a}$ and $\bm{\tilde{V}^a}$ for each description coding cost prediction, which are the approximations of $\bm{V^a}$ and $\bm{V^b}$, then $R(\bm{\tilde{V}^a})$ and $R(\bm{\tilde{V}^b})$ can be differentiable. As a result, the loss from context-based entropy estimation neural networks can be back-propagated to the multiple description encoder. Finally, the multiple description compressive loss for our trainable framework in Fig. \ref{Fig1} can be written in Eq. (\ref{mdobj}), in which $D_{r}$ is the parameter regularization term for our artificial neural networks. $\alpha$, $\beta$, and $\gamma$ are three hyperparameters. The $\bm{Y^a}$, $\bm{Y^b}$, and $\bm{Y}$ in the final multiple description compressive loss in Eq. (\ref{mdobj}) include $\bm{V^a}$ and $\bm{V^b}$ from the hard quantizers. To make it trainable, the solution of training our deep multiple description coding framework is operated like the Fig. \ref{DNNMODELa}, such that the forward-propagation uses hard quantization, but the back-propagation employs soft quantization.
\begin{figure*}[!htb]
\begin{align}
&f(\bm{X}, \bm{V^a}, \bm{V^b}, \bm{\tilde{V}^a}, \bm{\tilde{V}^b}, \bm{Y^a}, \bm{Y^b}, \bm{Y})=\gamma \cdot [R(\bm{\tilde{V}^a})+R(\bm{\tilde{V}^b})] + D(\bm{X},\bm{V^a}, \bm{V^b}, \bm{Y^a},\bm{Y^b},\bm{Y}) + \alpha \cdot D_{d} \notag\\
&=\gamma \cdot [R(\bm{\tilde{V}^a})+R(\bm{\tilde{V}^b})]+ [D_{1}(\bm{X},\bm{Y^a},\bm{Y^b},\bm{Y})+ D_{2}(\bm{X}, \bm{Y^a},\bm{Y^b}, \bm{Y})+ \beta \cdot D_{r}]+\alpha \cdot D_{d}
\label{mdobj}
\tag{13}
\end{align}
\hrulefill 
\end{figure*}

\begin{figure*}[!htb]
\centering
\includegraphics[width=4.3in]{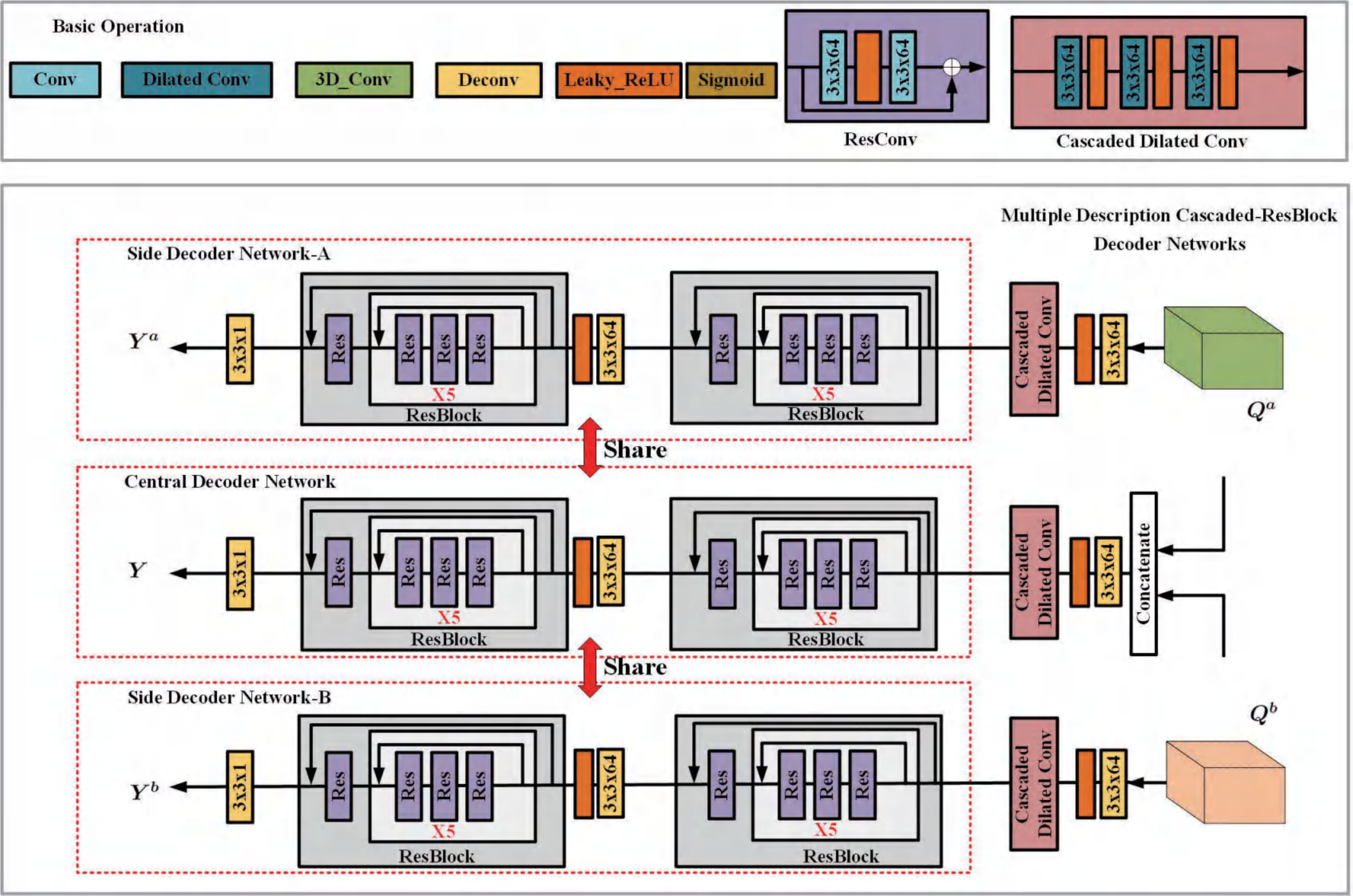}
\caption{The structure diagram of multiple description cascaded-ResBlock decoder networks with the parameter sharing.}
\label{Fig2}
\end{figure*}

\subsection{Network}
To fully explore image context information, we propose a multiple description multi-scale-dilated encoder network to create a feature tensor $\bm{Z}$, as well as two importance-indicator maps, $\bm{D^a}$ and $\bm{D^b}$, for multiple description generation, as shown in Fig. \ref{Fig1}. As discussed in \cite{qq23}, the dilated convolution can significantly enlarge the image receptive field, but it may introduce the grid effect. In \cite{qq24}, several dilated-convolutional layers are defined as a hybrid dilated convolution to resolve this problem for semantic segmentation. Inspired by these works, we use three cascaded dilated convolutions to extract multi-scale features since this is vital to leverage image context information for diversified multiple description generation. Each cascaded dilated convolution is composed of three layers of dilated convolutions, which is displayed in Fig. \ref{MDC-q-network}. First, a convolutional layer is used to transform the input image into 64-channel feature maps before operating three cascaded dilated convolutions. Second, each cascaded dilated convolution is followed by a $5\times5$ downsampling convolutional layer with a stride of 2 to shrink spatial size of the feature maps. Meanwhile, the first cascaded dilated convolution is followed by a downsampling convolutional layer with a stride of 4. Additionally, a convolutional layer with a stride of 2 is used to down-sample the output features of the second cascaded dilated convolution in the spatial domain. Finally, various features from different scales are concatenated and then aggregated together by a $3\times3$ convolutional layer to leverage image multi-scale context information, as depicted in Fig. \ref{MDC-q-network}. Although the feature tensor, as well as the importance-indicator maps can be separately generated by three structure-similar networks in the MD multi-scale-dilated encoder network, the input is shared by these networks to produce these tensors. If there is no parameter sharing, then the number of parameters is enormously increased, which results in consumption of memory and additional computational complexity. Accordingly, almost all of the convolutional layers can be shared in the MD multi-scale-dilated encoder network except for the last layers, to create different kinds of tensors (See the MD multi-scale-dilated encoder network in Fig. \ref{MDC-q-network}).

After receiving the de-quantized tensors $\bm{Q^a}$ and $\bm{Q^b}$, we propose to use MD cascaded-ResBlock decoder networks to decompress these tensors since the learning of ResConv (denoted as Res) with a shortcut connection is a type of residual learning, whose gradients can be easily back-propagated. The use of ResConv can also avoid the gradient vanishing problem. From Fig. \ref{MDC-q-network}, it can be seen that the side decoder networks and central decoder network share a similar network structure, except for different inputs. When all multiple descriptions are received, both de-quantized tensors $\bm{Q^a}$ and $\bm{Q^b}$ are concatenated and fed into the central decoder network for image decoding. If one description is missing but the other description is received, side decoder network-A or side decoder network-B is chosen to decode the de-quantized tensor $\bm{Q^a}$ or $\bm{Q^b}$. These networks are composed of three deconvolutional layers and two ResBlocks. Each of the first two deconvolutional layers is followed by one ResBlock. As depicted in Fig. \ref{MDC-q-network}, 16 ResConv blocks are employed with skip-connection in each ResBlock.

To overwhelmingly reduce the total number of parameters of the MD cascaded-ResBlock decoder networks, a symmetrical structure is designed. This structure includes two parts: the back layers in the decoder networks and the preceding convolutional layers in the encoder network, which share the parameters. The benefits of the parameter sharing include a decrease in the number of network parameters, the acceleration of training, and avoidance of over-fitting. However, there is a new problem that must be considered; that is, the parameter sharing will strongly affect the image coding efficiency, so we add a three-layer cascaded dilated convolution before parameter sharing, as shown in Fig. \ref{Fig2}, which will be ablated in the experimental section. As described above, the compact multiple description generation should be restricted by the entropy regularization term. To efficiently estimate the entropy of the feature tensor, M. Fabian and A. Eirikur, as well as co-authors presented context-based entropy estimation neural networks \cite{qq22}. Following the work of \cite{qq22}, we use two entropy estimation networks with a shortcut connection structure \cite{qq22}, which has six 3D-convolutional layers, as shown in Fig. \ref{MDC-q-network}. In this structure, the middle four convolutional layers are cascaded with two skip-connection, which construct two 3D-Resconv block.
\begin{table}[!htb]
\small
\centering
{
\caption{The setting of our framework and its variants. (Y=yes, N=no)}

\label{tableSQ}
\scriptsize
\begin{tabular}{|c|c|c|c|}
\hline
\multicolumn{ 1}{|c|}{Method} & \multicolumn{ 1}{|c|}{Distance } & \multicolumn{ 1}{|c|}{Using The Importance-} & \multicolumn{ 1}{|c|}{Parameter Sharing} \\

\multicolumn{ 1}{|c|}{/Conditions} & \multicolumn{ 1}{|c|}{Loss} & \multicolumn{ 1}{|c|}{Indicator Maps} & \multicolumn{ 1}{|c|}{for Decoder Networks} \\
\hline
   Ours-mr &    MR-SSIM &          Y &          N \\

   Ours-ms &    MS-SSIM &          Y &          N \\

Ours-mr-w/o &    MR-SSIM &          N &          N \\

Ours-mr(share) &    MR-SSIM &          Y &          Y \\
\hline
\end{tabular}
}
\end{table}
\section{Experimental results and analysis}
We evaluate the proposed method with respect to the structural similarity of the compressed images of several commonly available datasets (including Set4\footnote{\url{https://github.com/mdcnn/MDCNN_test40/tree/master/SET4}} used in \cite{qq11}, McMaster\footnote{\url{http://www4.comp.polyu.edu.hk/~cslzhang/CDM_Dataset.htm}}, Kodak PhotoCD dataset, denoted as Kodak\footnote{\url{http://www.r0k.us/graphics/kodak/}}, as well as SunHays\footnote{\url{https://github.com/jbhuang0604/SelfExSR/blob/master/README.md}}) . By default, two importance indicators are used for all of the models in the proposed framework, which also employs parameter sharing structure for the multiple description multi-scale-dilated encoder network. However, one model without the importance indicators will be clearly specified. When our framework uses MR-SSIM in the reconstruction loss and multiple description distance loss, the proposed method is marked as "Ours-mr". In contrast, the proposed method is denoted as "Ours-ms", if MS-SSIM is used for the reconstruction loss and multiple description distance loss. The "Ours-mr" model is labeled as "Ours-mr-w/o" when there is no parameter sharing for the multiple description cascaded-ResBlock decoder networks. When a model has the same settings as Ours-mr(share) but it has a symmetrical parameter sharing structure, then the final full-model of our framework is denoted as "Ours-mr(share)". To clearly display the differences between these models, they are listed in Table. \ref{tableSQ}.

As described in \cite{qq21}, SSIM is a good approximation to assess image quality from perspective of human visual perception, but this method only considers single-scale image information. Compared with SSIM, MS-SSIM is an image synthesis approach for image quality assessment that considers the relative importance of distorted images across different scales. Consequently, both MS-SSIM and MR-SSIM are chosen as objective measurements to assess distorted image quality, in addition to SSIM. Note that each scale SSIM weight factor of MR-SSIM is proportional to the image size, but MS-SSIM's weights are obtained according to visual testing. To demonstrate the coding efficiency of our MDC framework, our method is compared with several state-of-the-art MDC approaches, including the multiple description coding approach with randomly offset quantizers and the newest convolutional neural network-based standard-compatible method \cite{qq11} in terms of image coding efficiency when testing on several datasets. At last, visual comparisons of different MDC methods are provided to observe the image quality because human eyes are the ultimate recipients of the compressed images.

\begin{figure}[t]
\centering
\includegraphics[width=3.5in]{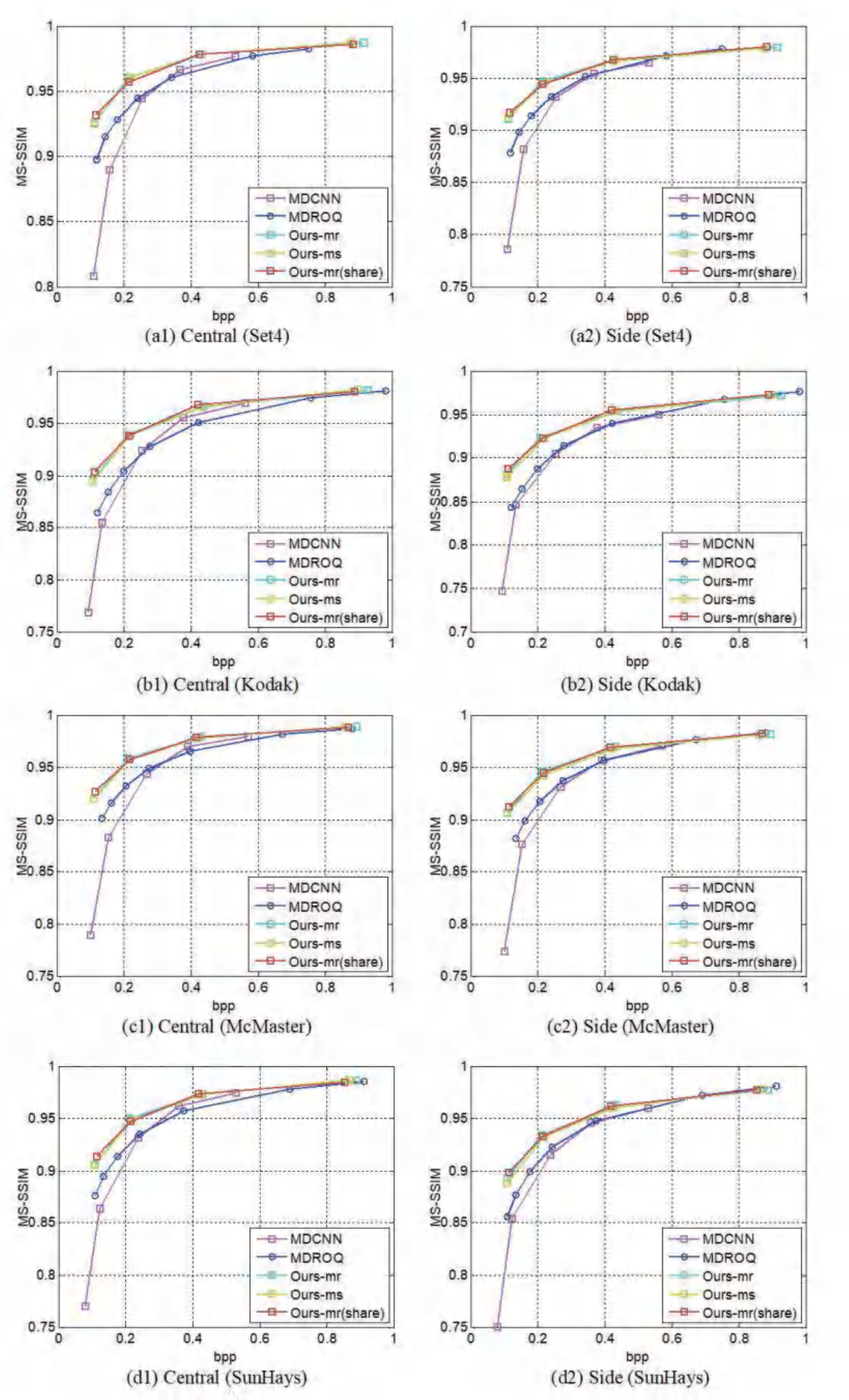}
\caption{The average objective quality comparisons of different multiple description coding methods. (a1-d1) are the MS-SSIM measurements for the decoded central images testing on Set4, Kodak, McMaster, and SunHays respectively. (a2-d2) are the MS-SSIM measurements for the decoded side images testing on Set4, Kodak, McMaster, and SunHays respectively.}
\label{ms-ssim}
\end{figure}

\begin{figure}[t]
\centering
\includegraphics[width=3.5in]{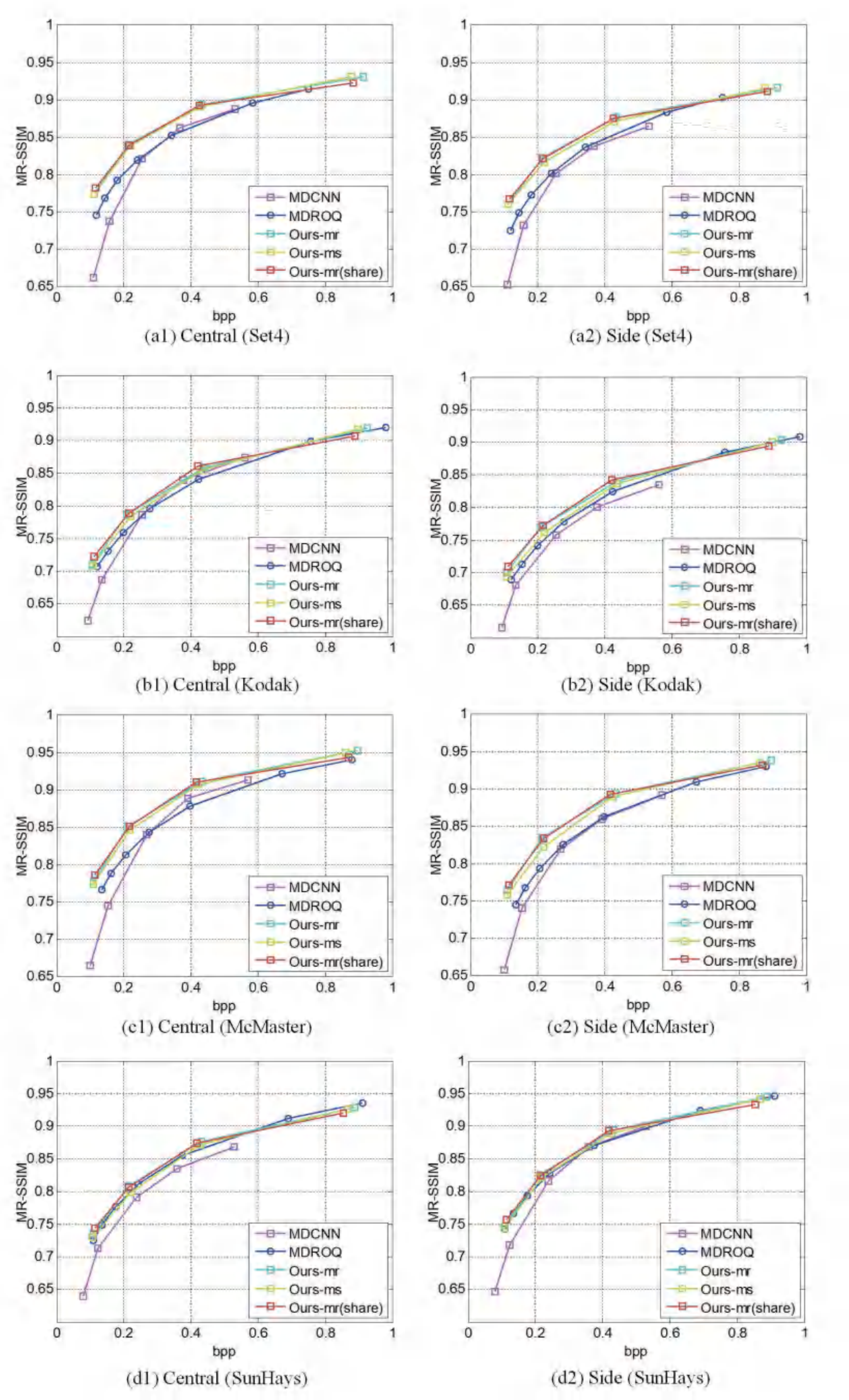}
\caption{The average objective quality comparisons of different multiple description coding methods. (a1-d1) are the MR-SSIM measurements for the decoded central images testing on Set4, Kodak, McMaster, and SunHays respectively. (a2-d2) are the MR-SSIM measurements for the decoded side images testing on Set4, Kodak, McMaster, and SunHays respectively.}
\label{mr-ssim}
\end{figure}

\subsection{Training details}
We train our framework on ImageNet training dataset from ILSVRC2012\footnote{\url{http://www.image-net.org/challenges/LSVRC/2012/}}. During training, each image patch with the size of $160 \times 160$ is obtained by randomly cropping the training images from this dataset. If the training image size is smaller than $160 \times 160$, we first resize the input image to be at least 160 in each dimension before cropping. Moreover, ImageNet's validation dataset is used as our validation dataset. Several commonly available datasets mentioned above are chosen as our testing datasets\footnote{\url{https://github.com/mdcnn/Deep-Multiple-Description-Coding}} for the comparison of different MDC methods. It should be noted that all of the testing image are cropped and/or resized to 16 integer multiples, which is required by one of the comparison methods. Among them, all image sizes of Set4 are $512\times512$, except for "Boat" image with a size of $256\times256$. The sizes of the datasets of McMaster, Kodak, and SunHays are, respectively, $496\times496$, $768\times512$ and $1024\times672$. During training, we choose Adam optimization to minimize the objective loss of our MDC framework with the initial learning rate of 4e-3 for our autoencoder network. The training batch size is set to 8, while the hyper parameter $\alpha$, $\beta$, and $\gamma$ are 0.1, 2e-4, and 0.1, respectively.

\begin{figure}[!htb]
\centering
\includegraphics[width=3.1in]{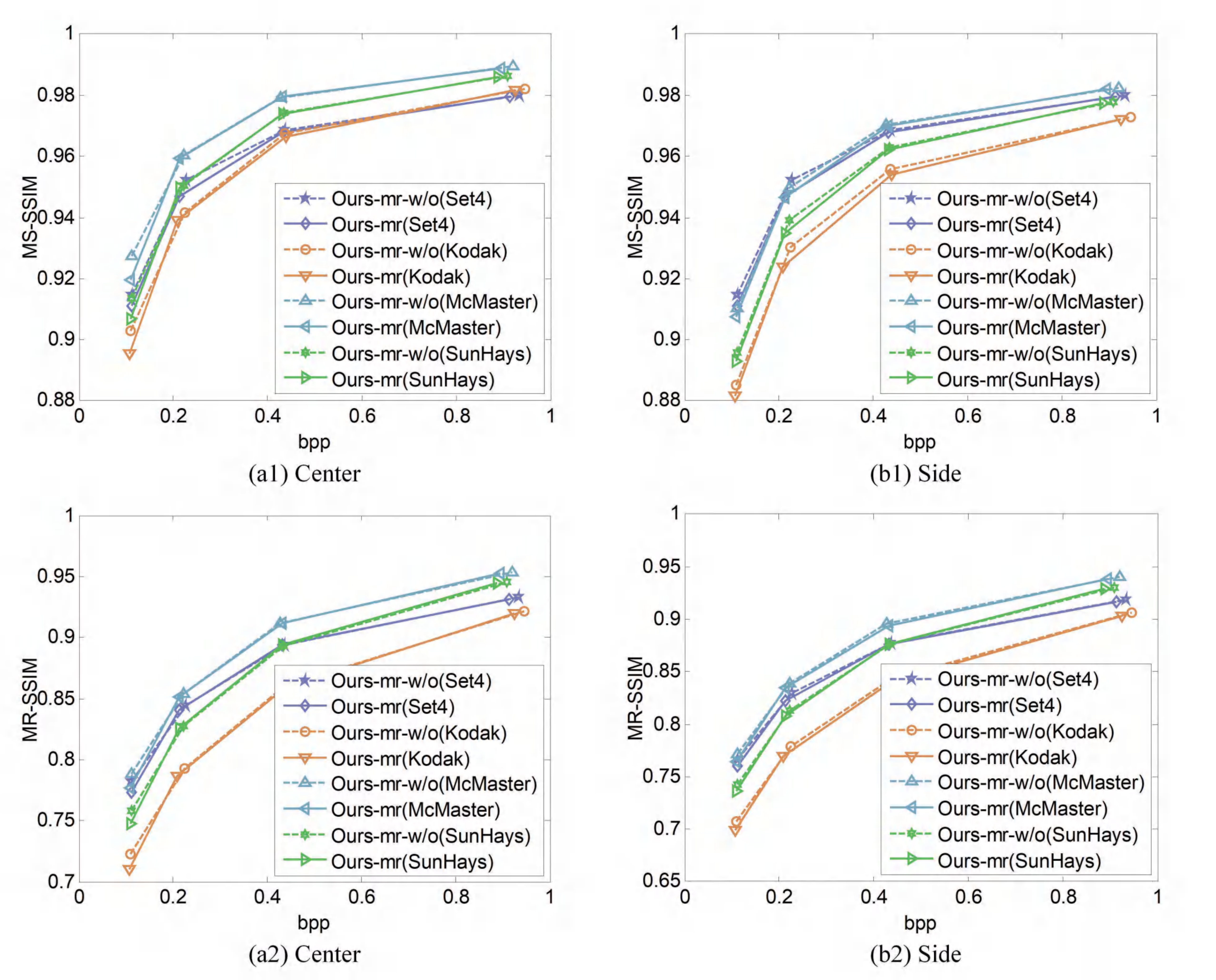}
\caption{The average objective quality comparisons between Ours-mr and Ours-mr-w/o testing on several datasets.}
\label{indicator-obj}
\end{figure}
\begin{figure}[!htb]
\centering
\includegraphics[width=3.1in]{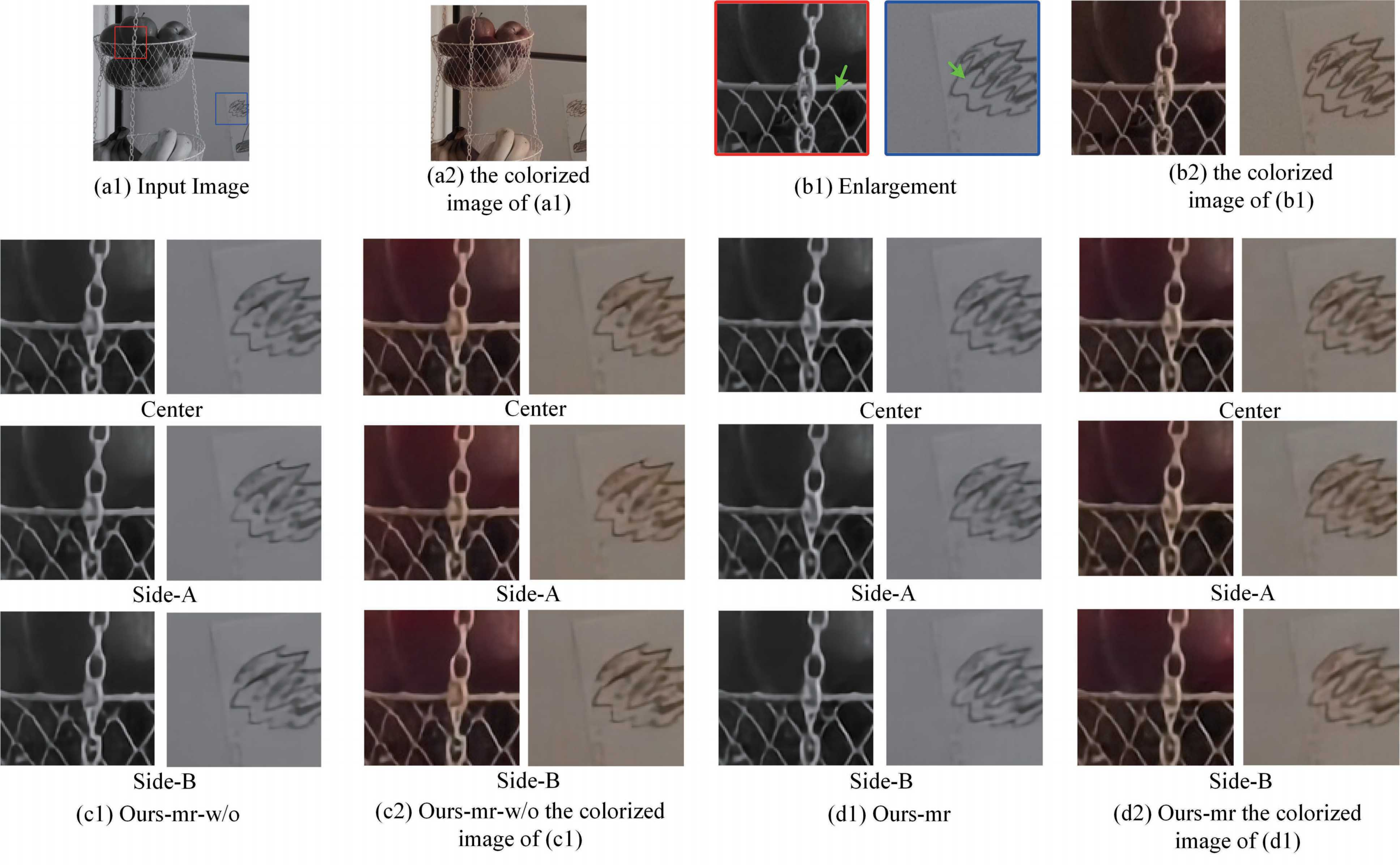}
\caption{The visual quality comparisons (b1-d1) between Ours-mr (0.39bpp) and Ours-mr-w/o (0.38bpp) and the colorized image comparisons (d2-d2) for their the decoded central and side images. (a1) is an image from McMaster, (a2) is the colorized image of (a1).}
\label{indicator}
\end{figure}
\subsection{Ablation studies}
As described above, the weights of MR-SSIM and MS-SSIM exert different effects on the MD coding efficiency, so the comparison between "Ours-mr" and "Ours-ms" is first given and discussed in the following. Fig. \ref{ms-ssim} and Fig. \ref{mr-ssim} show the objective performance comparison between "Ours-mr" and "Ours-ms" regarding the MR-SSIM and MS-SSIM measurements. Although the "Ours-mr" model is trained with MR-SSIM for the reconstruction loss and multiple description distance loss, this model always performs better than the "Ours-ms" model in terms of MR-SSIM and MS-SSIM when testing on the datasets of Set4, McMaster, Kodak, and SunHays. Accordingly, the other models such as "Ours-mr-w/o" and "Ours-mr(share)" are trained by using MR-SSIM in the proposed multiple description compressive loss.

\begin{figure}[!htb]
\centering
\includegraphics[width=3.5in]{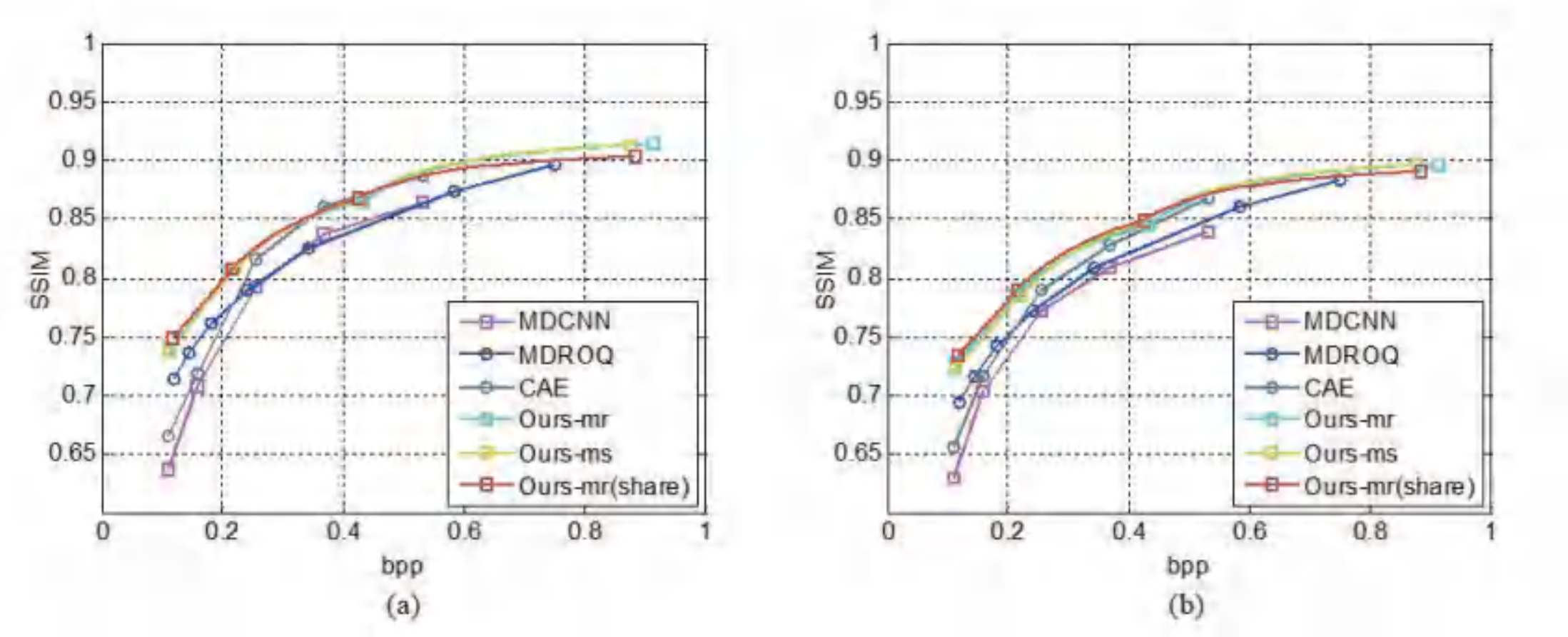}
\caption{The average objective quality comparisons of different multiple description coding methods. (a) shows the SSIM measurements for the decoded central images testing on Set4, (b) shows the SSIM measurements for the decoded side images testing on Set4.}
\label{CAE}
\end{figure}

\begin{figure}[t]
\centering
\includegraphics[width=2.5in]{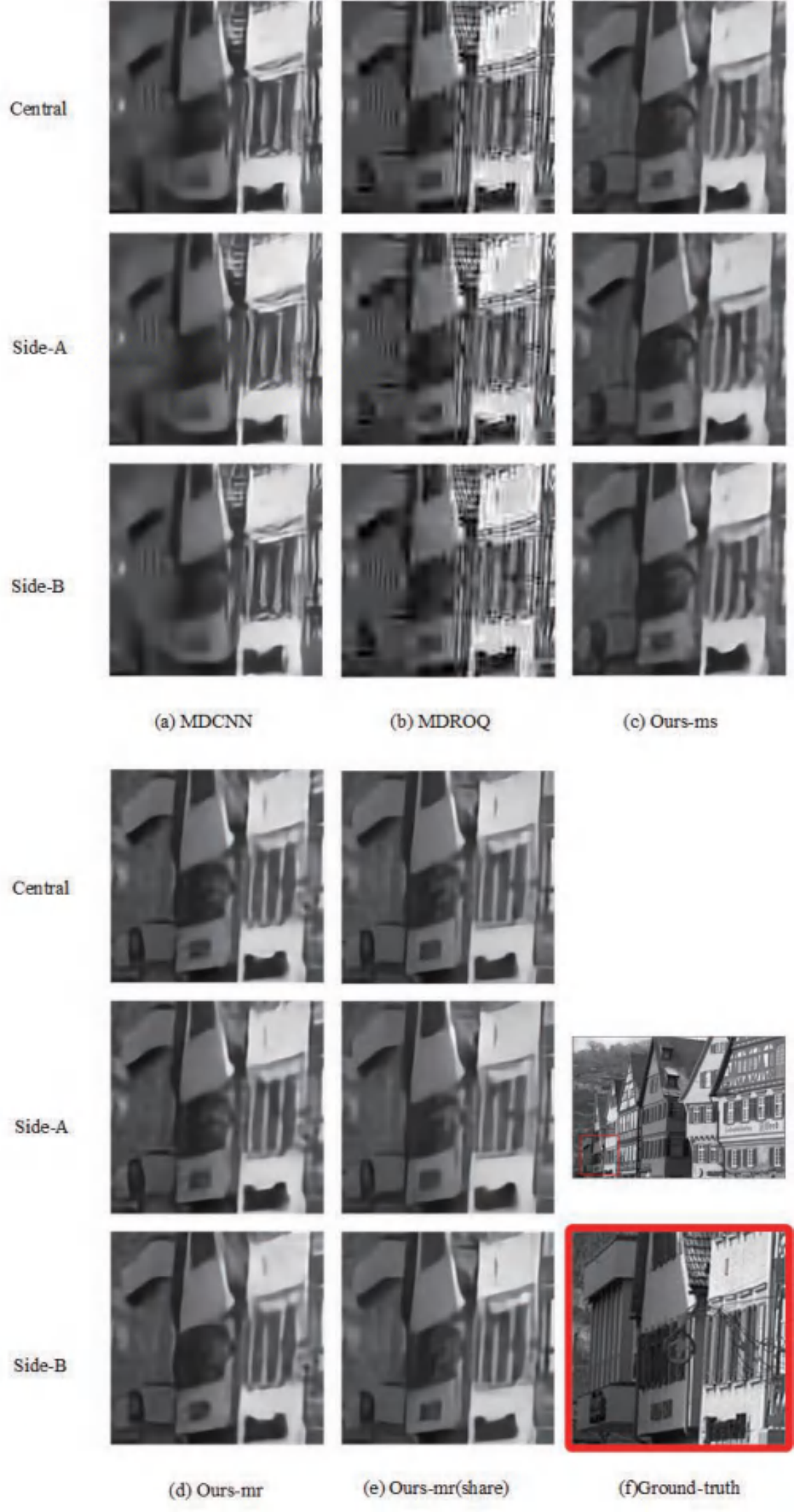}
\caption{The visual quality comparisons (a-e) of different multiple description coding methods with MDCNN (0.25bpp), MDROQ (0.29bpp), Ours-ms (0.29bpp), Ours-mr (0.25bpp), Ours-mr(share) (0.25bpp), (f) is an image from Kodak and its close-ups.}
\label{viscomp08patch}
\end{figure}

\begin{table*}[!htb]
\small
\centering
{
\caption{The comparison of running time(S) for different MDC methods when testing on an image with a size of 512x512. }
\label{tabletimes}
\begin{tabular}{|c|c|c|c|c|c|}
\hline
    Method &      MDROQ \cite{qq13} &      MDCNN \cite{qq11} &    Ours-mr &    Ours-ms & Ours-mr(share) \\
\hline
Encoding Time &        0.8 &        0.1 &        0.6 &        0.5 &        0.5 \\
\hline
Center Decoding &        8.4 &       0.05 &        3.9 &        4.3 &        0.2 \\
\hline
Average Side Decoding &        0.6 &       0.03 &        1.1 &        1.2 &       0.07 \\
\hline
Coding Time &        9.8 &       0.18 &        5.6 &          6 &       0.77 \\
\hline
\end{tabular}

}
\end{table*}

{\bf When we encode an image with one importance-indicator map as like \cite{qq22, qqlimu}, a pair of quantizers is able to be learned to quantize the same tensor to generate different descriptions. Thus, in this case, the diversity of multiple description generation only depends on the quantizers, which affects image coding efficiency since the one importance-indicator map only controls the ROI coding, which contributes almost nothing to the diversity of multiple description tensors, thus affecting coding performance (The corresponding experimental results is illustrated in the supplementary material\footnote{\url{https://github.com/mdcnn/Deep-Multiple-Description-Coding}}).}

In our deep MDC framework, each quantizer is accompanied by an importance-indicator map to generate the diversified multiple descriptions. To see the significance of the importance-indicator maps in the proposed framework, we compare the proposed models "Ours-mr" and "Ours-mr-w/o" in Fig. \ref{indicator-obj} (a1-d1), from which it can be found that the objective performances of these models are very similar. But, the decoded side images and central images compressed by "Ours-mr" and "Ours-mr-w/o" exhibit some differences on the structural preservation in terms of image spatial changes, which can be seen in Fig. \ref{indicator}. To better observe the performance of these models, whose compressed images are colorized by "Let there be Color!" \cite{qqc2}, as shown in Fig. \ref{indicator} (a1-d1) and (a2-d2). This figure shows that the "Ours-mr" model can retain more image spatial structures than "Ours-mr" after multiple description coding (see the areas to which the green arrows point in Fig. \ref{indicator} (b1)).

Since the trained model's size not only affects each model's computational complexity, but also restricts the applications of the proposed framework, the network parameter sharing in the proposed framework is studied. First, we investigate how much the performance of the proposed framework is influenced by the network parameter sharing. The figures of Fig. \ref{ms-ssim} and Fig. \ref{mr-ssim} display the objective comparison of "Ours-mr" and "Ours-mr(share)". From these figures, we can conclude that the performance of "Ours-mr(share)" is very close to that of "Ours-mr" in most cases. However, the objective measurement of the "Ours-mr(share)" model is slightly lower than that of the "Ours-mr" model at very high bit rates. Meanwhile, "Ours-mr(share)" with a symmetrical structure for the parameter sharing, can reduce the number of parameters to approximately 0.436 times that of the "Ours-mr" model. The number of network parameter of "Ours-mr(share)" is 0.406 times that of the model for the proposed framework without any parameter sharing. From these results, it can be deduced that the parameter sharing with the symmetrical structure can greatly reduce the number of the model parameters in the proposed framework.

\subsection{Objective and visual quality comparisons of different methods}
To validate the efficiency of the proposed framework, we compare our method with the latest standard-compatible CNN-based MDC method \cite{qq11}, with convolutional auto-encoder-based multiple description coding, and with a multiple description coding approach with randomly offset quantizers \cite{qq13}, which are denoted as "MDCNN" \cite{qq11}, "CAE" \cite{qq11}, and "MDROQ" \cite{qq13} respectively. {\bf As shown in Fig. \ref{CAE}, our method has higher coding efficiency than the methods of "MDCNN" \cite{qq11}, "CAE" \cite{access}, and "MDROQ" \cite{qq13} at all times in terms of SSIM, when testing on the Set4 dataset.}

Although "Ours-mr" is trained with MR-SSIM instead of MS-SSIM, both the MR-SSIM and MS-SSIM results of several comparative MDC approaches are provided in Fig. \ref{ms-ssim} and Fig. \ref{mr-ssim}, after testing on several datasets. From these figures, we can observe that "Ours-mr" has better performance than "Ours-ms" regarding MR-SSIM and MS-SSIM. Moreover, the objective MS-SSIM measurements of the side decoded images between "MDCNN" and "MDROQ" are very similar and the objective MS-SSIM when the bit rate is higher than approximately 0.3 bpp when testing on the Set4, Kodak, McMaster, and SunHays datasets. The MR-SSIM measurements of the central decoded images compressed by "MDCNN" can compete with or even exceed those of "MDROQ". However, "MDROQ" has exhibits better performance than "MDCNN" in terms of MS-SSIM and MR-SSIM at very low bit rates. "Ours-mr" and "Ours-mr(share)", as well as "Ours-ms" have the best coding efficiencies on all testing datasets with respect to both side and central decoded images compared with "MDCNN" and "MDROQ" in terms of MR-SSIM and MS-SSIM, as depicted in Fig. \ref{ms-ssim} and Fig. \ref{mr-ssim}. From these figures, we can see that the coding efficiencies of "Ours-mr", as well as "Ours-mr(share)" and "Ours-ms" are far higher than those of the comparative methods at low bit rates when testing on several publicly available datasets. Meanwhile, the MR-SSIM and MS-SSIM of "Ours-mr" and "Ours-mr(share)" are superior to those of "Ours-ms".

"Ours-mr", "Ours-mr(share)" and "Ours-ms" can retain more structures of each object than "MDCNN" \cite{qq11} and "MDROQ" \cite{qq13} for both side and central decoded images, as displayed in Fig. \ref{viscomp08patch} (a-e). Although "MDCNN" \cite{qq11} can preserve some small structures, this method makes many significant objects disappear. At the same time, some objects are enormously distorted when testing images are compressed by "MDCNN" \cite{qq11} and "MDROQ" \cite{qq13}, which can be seen in Fig. \ref{viscomp08patch} (a-b). Moreover, "Ours-mr" as well as "Ours-mr(share)" and "Ours-ms" does not contain obvious visual noises, such as coding artifacts, compared to "MDROQ" \cite{qq13}, which can be seen in Fig. \ref{viscomp08patch} (c-e), while our side and central decoded images appear to more natural.

\subsection{Comparisons of coding time}
{\bf TABLE \ref{tabletimes} provides the encoding time and decoding time of several MDC methods for comparison when testing on an image with a size of 512x512. In this table, all the operations of convolutional neural networks for these MDC methods are run on an NVIDIA-GTX1080 GPU device. Among these comparative MDC methods in TABLE \ref{tabletimes}, "MDCNN" requires the least coding time. From this table, it can be observed that "Ours-mr" and "Ours-ms" have very similar performances for image encoding, center decoding and average side decoding. The running time of these two models for encoding as well as decoding are greater than that of "Ours-mr(share)". Compared to "Ours-mr", "Ours-ms", "Ours-mr(share)" and "MDCNN", "MDROQ" requires the most time for image coding. }

\section{Conclusion}
In this paper, we propose a deep multiple description coding framework in which MD quantizers are automatically learned in an end-to-end manner during training. Meanwhile, a symmetrical parameter sharing structure is designed for our autoencoder networks to overwhelmingly reduce the total number of neural network parameters. Ablation studies on whether two importance-indictor maps are essential or not, how to control the redundancy between different descriptions and different kinds of structural dissimilarity loss functions, as well as parameter sharing, are provided in the section of experimental results and analysis. At last, we demonstrate that our method offers better coding efficiency than several advanced MD image compression methods, when tested on commonly available datasets, especially at low bit rates.




\bibliographystyle{IEEEtran}
\bibliography{IEEEfull,MDQNN}

\begin{thebibliography}{10}
\providecommand{\url}[1]{#1}
\csname url@samestyle\endcsname
\providecommand{\newblock}{\relax}
\providecommand{\bibinfo}[2]{#2}
\providecommand{\BIBentrySTDinterwordspacing}{\spaceskip=0pt\relax}
\providecommand{\BIBentryALTinterwordstretchfactor}{4}
\providecommand{\BIBentryALTinterwordspacing}{\spaceskip=\fontdimen2\font plus
\BIBentryALTinterwordstretchfactor\fontdimen3\font minus
  \fontdimen4\font\relax}
\providecommand{\BIBforeignlanguage}[2]{{%
\expandafter\ifx\csname l@#1\endcsname\relax
\typeout{** WARNING: IEEEtran.bst: No hyphenation pattern has been}%
\typeout{** loaded for the language `#1'. Using the pattern for}%
\typeout{** the default language instead.}%
\else
\language=\csname l@#1\endcsname
\fi
#2}}
\providecommand{\BIBdecl}{\relax}
\BIBdecl

\bibitem{qq2}
M.~Kazemi, R.~Iqbal, and S.~Shirmohammadi., ``{Joint intra and multiple
  description coding for packet loss resilient video transmission},''
  \emph{IEEE Transactions on Multimedia}, vol.~20, no.~4, pp. 781--795, 2018.

\bibitem{qq11}
L.~Zhao, H.~Bai, A.~Wang, and Y.~Zhao, ``{Multiple description convolutional
  neural networks for image compression},'' \emph{IEEE Transactions on Circuits
  and Systems for Video Technology}, vol.~29, no.~8, pp. 2494--2508, 2019.

\bibitem{qq12}
Y.~Xu and C.~Zhu, ``{End-to-end rate-distortion optimized description
  generation for H. 264 multiple description video coding},'' \emph{IEEE
  Transactions on Circuits and Systems for Video Technology}, vol.~23, no.~9,
  pp. 1523--1536, 2013.

\bibitem{qq4}
F.~Shirani and S.~S. Pradhan, ``{An achievable rate-distortion region for
  multiple descriptions source coding based on coset codes},'' \emph{IEEE
  Transactions on Information Theory}, vol.~64, no.~5, pp. 3781--3809, 2018.

\bibitem{qq27}
V.~A. Vaishampayan, ``{Design of multiple description scalar quantizers},''
  \emph{IEEE Transactions on Information Theory}, vol.~39, no.~3, pp. 821--834,
  1993.

\bibitem{qq28}
V.~K. Goyal, ``{Scalar quantization with random thresholds},'' \emph{IEEE
  Signal Processing Letters}, vol.~18, no.~9, pp. 525--528, 2011.

\bibitem{qq35}
Y.~Wang, M.~T. Orchard, V.~Vaishampayan, and A.~R. Reibman, ``{Multiple
  description coding using pairwise correlating transforms},'' \emph{IEEE
  Transactions on Image Processing}, vol.~10, no.~3, pp. 351--366, 2001.

\bibitem{qq36}
V.~K. Goyal and J.~Kovacevic, ``{Generalized multiple description coding with
  correlating transforms},'' \emph{IEEE Transactions on Information Theory},
  vol.~47, no.~6, pp. 2199--2224, 2001.

\bibitem{qq37}
D.~Saitoh and T.~Yakoh, ``{Ratio configurable multiple description correlating
  transforms coding},'' in \emph{IEEE International Conference on Industrial
  Technology}, Auburn, Mar. 2011.

\bibitem{qq3}
M.~Majid, M.~Owais, and S.~M. Anwar, ``{Visual saliency based redundancy
  allocation in HEVC compatible multiple description video coding},''
  \emph{Multimedia Tools and Applications}, vol.~77, no.~16, pp.
  20\,955--20\,977, 2018.

\bibitem{qq19}
L.~Zhao, H.~Bai, A.~Wang, and Y.~Zhao, ``{Learning a virtual codec based on
  deep convolutional neural network to compress image},'' \emph{Journal of
  Visual Communication and Image Representation}, vol.~63, no.~1, pp.
  102\,589--102\,599, 2019.

\bibitem{qq17}
H.~Wu, T.~Zheng, and S.~Dumitrescu, ``{On the design of symmetric
  entropy-constrained multiple description scalar quantizer with linear joint
  decoders},'' \emph{IEEE Transactions on Communications}, vol.~65, no.~8, pp.
  3453--3466, 2017.

\bibitem{qq13}
L.~Meng, J.~Liang, U.~Samarawickrama, Y.~Zhao, H.~Bai, and A.~Kaup, ``Multiple
  description coding with randomly and uniformly offset quantizers,''
  \emph{IEEE Transactions on Image Processing}, vol.~23, no.~2, pp. 582--95,
  2014.

\bibitem{qq10}
S.~Dumitrescu and Y.~Wan, ``{Bit-error resilient index assignment for multiple
  description scalar quantizers},'' \emph{IEEE Transactions on Information
  Theory}, vol.~61, no.~5, pp. 2748--2763, 2015.

\bibitem{qq26}
H.~Jafarkhani and V.~Tarokh, ``{Multiple description trellis-coded
  quantization},'' \emph{IEEE Transactions on Communications}, vol.~47, no.~6,
  pp. 799--803, 1999.

\bibitem{qq14}
V.~Vaishampayan, N.~Sloane, and S.~Servetto, ``Multiple-description vector
  quantization with lattice codebooks: design and analysis,'' \emph{IEEE
  Transactions on Information Theory}, vol.~47, no.~5, pp. 1718--1734, 2001.

\bibitem{qq38}
G.~Romano, P.~S. Rossi, and F.~Palmieri, ``{Multiple description image coder
  using correlating transforms},'' in \emph{European Signal Processing
  Conference}, Vienna, 2015.

\bibitem{qq8}
J.~Chen, C.~Cai, L.~Li, and C.~Li, ``{Layered multiple description video coding
  using dual-tree discrete wavelet transform and H. 264/AVC},''
  \emph{Multimedia Tools and Applications}, vol.~75, no.~5, pp. 2801--2814,
  2016.

\bibitem{qq15}
V.~Goyal, J.~Kelner, and J.~Kovacevic, ``Multiple description vector
  quantization with a coarse lattice,'' \emph{IEEE Transactions on Information
  Theory}, vol.~48, no.~3, pp. 781--788, 2002.

\bibitem{qq18}
S.~Dumitrescu, Y.~Chen, and J.~Chen, ``{Index mapping for bit-error resilient
  multiple description lattice vector quantizer},'' \emph{IEEE Transactions on
  Communications}, vol.~PP, no.~99, pp. 1--1, 2018.

\bibitem{qq7}
Z.~Gao and S.~Dumitrescu, ``Flexible multiple description lattice vector
  quantizer with $l\geq 3$ descriptions,'' \emph{IEEE Transactions on
  Communications}, vol.~62, no.~12, pp. 4281--4292, 2014.

\bibitem{qq30}
N.~Franchi, M.~Fumagalli, R.~Lancini, and S.~Tubaro, ``{Multiple description
  video coding for scalable and robust transmission over IP},'' \emph{IEEE
  Transactions on Circuits and Systems for Video Technology}, vol.~15, no.~3,
  pp. 321--334, 2005.

\bibitem{qq29}
N.~Gadgil, H.~Li, and E.~J. Delp, ``{Spatial subsampling-based multiple
  description video coding with adaptive temporal-spatial error concealment},''
  in \emph{Picture Coding Symposium}, Cairns, May 2015.

\bibitem{qq31}
G.~Zhang and R.~L. Stevenson, ``{Efficient error recovery for multiple
  description video coding},'' in \emph{Picture Coding Symposium}, Paris, Oct.
  2004.

\bibitem{qq32}
C.~Zhu and M.~Liu, ``{Multiple description video coding based on hierarchical B
  pictures},'' \emph{IEEE Transactions on Circuits and Systems for Video
  Technology}, vol.~19, no.~4, pp. 511--521, 2009.

\bibitem{qqAR}
X.~Zhang, W.~Yang, Y.~Hu, and J.~Liu, ``{DMCNN: Dual-domain Multi-scale
  Convolutional Neural Network for Compression Artifacts Removal},'' in
  \emph{IEEE International Conference on Image Processing}, Athens, Oct. 2018.

\bibitem{qq41}
E.~Agustsson, F.~Mentzer, M.~Tschannen, L.~Cavigelli, R.~Timofte, L.~Benini,
  and L.~V. Gool, ``{Soft-to-hard vector quantization for end-to-end learned
  compression of images and neural networks},'' in \emph{Advances in Neural
  Information Processing Systems}, California, Dec. 2017.

\bibitem{qq22}
M.~Fabian, A.~Eirikur, T.~Michael, T.~Radu, and V.~G. Luc, ``{Conditional
  probability models for deep image compression},'' in \emph{IEEE Conference on
  Computer Vision and Pattern Recognition}, Salt Lake City, Jun. 2018.

\bibitem{qq40}
L.~Theis, W.~Shi, A.~Cunningham, and F.~Huszar, ``{Lossy image compression with
  compressive autoencoders},'' in \emph{International Conference on Learning
  Representations}, Palais, Apr. 2017.

\bibitem{qq42}
G.~Toderici, S.~Malley, S.~Hwang, D.~Vincent, D.~Minnen, S.~Baluja, and et~al.,
  ``{Variable rate image compression with recurrent neural networks},'' in
  \emph{International Conference on Learning Representations (ICLR)}, Puerto
  Rico, May 2016.

\bibitem{vggloss}
O.~Rippel and L.~Bourdev, ``{Real-time adaptive image compression},'' in
  \emph{International Conference on Machine Learning}, Sydney, Aug. 2017.

\bibitem{access}
H.~Li, L.~Meng, J.~Zhang, Y.~Tan, Y.~Ren, and H.~Zhang, ``Multiple description
  coding based on convolutional auto-encoder,'' \emph{IEEE Access}, vol.~7,
  no.~1, pp. 26\,013--26\,021, 2019.

\bibitem{qqroi}
A.~Jerbi, W.~Jian, and S.~Shirani, ``{Error-resilient region-of-interest video
  coding},'' \emph{IEEE Transactions on Circuits and Systems for Video
  Technology}, vol.~15, no.~9, pp. 1175--1181, 2005.

\bibitem{qqlimu}
M.~Li, W.~Zuo, S.~Gu, D.~Zhao, and D.~Zhang, ``{Learning convolutional networks
  for content-weighted image compression},'' in \emph{IEEE Conference on
  Computer Vision and Pattern Recognition}, Salt Lake City, Jun. 2018.

\bibitem{qq20}
B.~Yochai and M.~Tomer, ``{The perception-distortion tradeoff},'' in \emph{IEEE
  Conference on Computer Vision and Pattern Recognition}, Salt Lake City, Jun.
  2018.

\bibitem{pr2019}
L.~Zhao, H.~Bai, J.~Liang, B.~Zeng, A.~Wang, and Y.~Zhao, ``Simultaneous
  color-depth super-resolution with conditional generative adversarial
  networks,'' \emph{Pattern Recognition}, vol.~88, no.~1, pp. 356--369, 2019.

\bibitem{pix2pix}
P.~Isola, J.~Y. Zhu, T.~Zhou, and A.~A. Efros, ``{Image-to-image translation
  with conditional adversarial networks},'' in \emph{IEEE Conference on
  Computer Vision and Pattern Recognition}, Honolulu, Jul. 2017.

\bibitem{qq21}
Z.~Wang, E.~P. Simoncelli, and A.~C. Bovik, ``{Multiscale structural similarity
  for image quality assessment},'' in \emph{The Thrity-Seventh Asilomar
  Conference on Signals, Systems and Computers, 2003}, Pacific Grove, Nov.
  2003.

\bibitem{qq23}
F.~Yu and V.~Koltun, ``{Multi-scale context aggregation by dilated
  convolutions},'' in \emph{arXiv:1511.07122}, 2015.

\bibitem{qq24}
P.~Wang, P.~Chen, D.~L. Y.~Yuan, Z.Huang, X.~Hou, and G.Cottrell,
  ``{Understanding convolution for semantic segmentation},'' in \emph{IEEE
  Winter Conference on Applications of Computer Vision}, Lake Tahoe, Mar. 2018.

\bibitem{qqc2}
S.~Iizuka, E.~Simoserra, and H.~Ishikawa, ``{Let there be color!: joint
  end-to-end learning of global and local image priors for automatic image
  colorization with simultaneous classification},'' \emph{{ACM Transactions on
  Graphics}}, vol.~35, no.~4, pp. 1--11, 2016.

\end{thebibliography}
\end{document}